\documentclass[journal]{IEEEtran}

\usepackage{tabularx,booktabs}
\usepackage[justification=centering]{caption}

\usepackage{ifpdf}
\ifpdf
\else
\fi

\usepackage{cite}
\ifCLASSINFOpdf
\usepackage[pdftex]{graphicx}
\else
\fi

\usepackage{amsmath}

\usepackage{algorithmic}
\usepackage{algorithm}

\usepackage{array} 
\usepackage{amsthm}
\newtheorem{myDef}{Definition}

\ifCLASSOPTIONcompsoc 
\usepackage{subfigure}
\usepackage[caption=false,font=normalsize,labelfont=sf,textfont=sf]{subfig}
\else
\usepackage[caption=false,font=footnotesize]{subfig}
\fi

\usepackage{url}

\makeatletter
\let\NAT@parse\undefined
\makeatother
\usepackage{hyperref}  

\begin{document}
%
\title{MMGET: A Markov model for generalized evidence theory}
%
%
%

\author{Yuanpeng~He,
	    Fuyuan Xiao
        
\thanks{Yuanpeng He: School of Computer and Information Science, Southwest University, Chongqing, 400715, China}
\thanks{
	Corresponding author: Fuyuan Xiao, School of
	Computer and Information Science, Southwest University, Chongqing,
	400715, China. Email address: xiaofuyuan@swu.edu.cn,
	doctorxiaofy@hotmail.com.}}

%
%

\markboth{Journal of \LaTeX\ Class Files,~Vol.~14, No.~8, August~2015}%
{Shell \MakeLowercase{\textit{$et$ $al.$}}: Bare Demo of IEEEtran.cls for IEEE Communications Society Journals}
%



\maketitle

\begin{abstract}
In real life, lots of information merges from time to time. To appropriately describe the actual situations, lots of theories have been proposed. Among them, Dempster-Shafer evidence theory is a very useful tool in managing uncertain information. To better adapt to complex situations of open world, a generalized evidence theory is designed. However, everything occurs in sequence and owns some underlying relationships with each other. In order to further embody the details of information and better conforms to situations of real world, a Markov model is introduced into the generalized evidence theory which helps extract complete information volume from evidence provided. Besides, some numerical examples is offered to verify the correctness and rationality of the proposed method. 
\end{abstract}

\begin{IEEEkeywords}
Dempster-Shafer evidence theory \ \ Open world\ \ Generalized evidence theory \ \ Markov model
\end{IEEEkeywords}
\section{Introduction}
The world is full of different kinds of information and more and more information is produced. Therefore, how to properly measure information with uncertainty has become a hot topic in recent years. Because the measure of information is very helpful to extract a general figure of information provided before a concise process of it. To satisfy this kind of need to appropriately measure information, lots of related theories have been designed. Among them, some theories are very representative in managing uncertain information, such as fuzzy mathematics \cite{Li2020generateTDBF, fei2019pythagorean, song2019divergencebased, zhanglimao2019}, the extension of evidence theory \cite{Xiao2020evidencecombination, Deng2020InformationVolume, xu2018belief, mao2020DEMATELevidence}, soft theories \cite{feng2016soft, Song2019POWA, tian2020zslf, Fei2019Evidence}, $D$- number \cite{IJISTUDNumbers, deng2019evaluating, liu2020anextended, LiubyDFMEA}, $Z$- number \cite{kang2019deriveknowledge, Jiang2019Znetwork} and maximum theory \cite{Xiao2020maximum, deng2020negationINS, gao2020pseudopascal, Yager2014}. All of the theories mentioned performs well in extracting truly useful information from uncertainty. With the development of technology of computer, some new visions on the presentation form of information has been proposed. The related works attempts to dispose information from a completely new dimension, which are quantum theory \cite{lai2020parrondo, dengentropyentanglement, Dai2020interferenceQLBN, Gao2019quantummodel} and complex function \cite{Xiao2020CEQD, Harish2019NovelAggregation, Xiao2019complexmassfunction}. Due to the effectiveness of these theories in disposing information, some practical applications also benefit from these meaningful works, such as target recognition \cite{Pan2020association, LiuF2020TFS}, decision making \cite{zhou2018evidential, fu2020multiple, fu2020comparison,liao2020deng} and pattern classification \cite{Xiao2019Distancemeasure, liu2020evidence}.

Among all of the previously proposed theories, Dempster-Shafer evidence theory (D-S evidence theory) \cite{Dempster1967Upper, book} is the representative work to handle uncertainty contained in information. To better adapt to actual environment of application in an open world, an improved version of D-S evidence theory is proposed, namely generalized evidence theory (GET) \cite{DBLP:journals/apin/Deng15}. However, the two theory consider all of the incidents as a still figure, which is lack of description on the dynamic process of the transition of different things. Fortunately, the Markov chain can be utilized to avoid this drawback by setting all the incidents under judgment at a random and mutually connected process. Some works have taken the Markov chain into different applications \cite{DBLP:journals/ivc/PieczynskiB06, DBLP:conf/ecsqaru/Soubaras09, DBLP:journals/fss/ZhuCG16,DBLP:journals/trob/Zhu91}, which works very well. Therefore, in order to enable the generalized evidence theory to have a dynamic figure on the evidences given, a specially customised Markov model is introduced into GET. So, when the Markov model is integrated into the GET, the completely new frame of evidence is able to possess some new properties which is given as follows:
\begin{enumerate}[(1)]
	\item The improved generalized evidence theory can manifest the process of transition of propositions, which describes a dynamic process of evidences.
	\item Transition probability and further extensions of it can be generated as a dual certificate of the degree of belief of the first dimension.
	\item Completely new distance measure, similarity measure, uncertainty measure and method of combination are proposed based on the concept of MMGET.
\end{enumerate}

The rest of paper is written as follows. The section of preliminary introduces some basic concepts related to the work proposed in this paper. Then, the next section present every details of the proposed model. And the part of numerical examples gives some examples to verify the validity and correctness of the proposed method.

\section{Preliminaries}
In this section, some basic concepts are introduced. And lots of meaningful work have been completed to solve different kinds of problems \cite{Luo2020negation, yanhy2020entropy,Zhao2020complex,song2020selfadaptive,Li2021Ageneralized}.
\subsection{Generalized evidence theory (GET) \cite{DBLP:journals/apin/Deng15}}
\begin{myDef}Mass function\end{myDef} Assume there exists a frame of discernment (FOD), $A$, in an open world. $2_{G}^{A}$ denotes the power set of the whole FOD which is consist of $2^{A}$ elements. For $\forall P \subset A$, a mass function which is also a mapping can be defined as:
\begin{equation}
	m_{G}: 2_{G}^{A} \rightarrow [0,1]
\end{equation}

And the properties the mass function satisfies can be defined as:
\begin{equation}
	\sum_{P \in 2^{A}_{G}} m_{G}(P) = 1
\end{equation}
\begin{equation}
	m_{G}(\emptyset) \geq 0
\end{equation}

And thus the $m_{G}$ is a generalized basic probability assignment (GBPA) of the FOD, $A$. Besides, it can be concluded that the $m_{G}(\emptyset) = 0$ is not a restriction in GBPA, which means the element $\emptyset$ can be also regarded as a focal element in the generalized evidence theory. When the value of $m_{G}(\emptyset)$ is equal to 0, the GBPA can exactly degenerate in to the form of classic BPA.

\begin{myDef}Generalized belief function (GBF)\end{myDef}
Assume there is a GBPA $a$, the GBF with respect to $a$ can be defined as:
\begin{equation}
	GBF(a) = \sum_{b \subseteq a} m(b)
\end{equation}
\begin{equation}
	GBF(\emptyset) = m(\emptyset)
\end{equation}

\begin{myDef}Generalized plausible function (GPF)\end{myDef}
Assume there is a GBPA $a$, the GPF with respect to $a$ can be defined as:
\begin{equation}
	GPF(a) = \sum_{b \cap a \neq \emptyset} m(b)
\end{equation}
\begin{equation}
	GPF(\emptyset) = m(\emptyset)
\end{equation}

\begin{myDef}Generalized combination rule (GCR)\end{myDef}
In the GET, given two GBPAs contained in the same frame of discernment, the GCR can be defined as:
\begin{equation}
	m(D) = \frac{(1 - m(\emptyset))\sum_{Q\cap W = D}m_{1}(Q)m_{2}(W)}{1 - K}
\end{equation}
\begin{equation}
	K = \sum_{Q \cap W = \emptyset}m_{1}(Q)m_{2}(W)
\end{equation}
\begin{equation}
	m(\emptyset) = m_{1}(\emptyset)m_{2}(\emptyset)
\end{equation}
\begin{equation}
	m(\emptyset) = 1\ \  if \ \ and \ \ only \ \ if \ \ K = 1
\end{equation}

\begin{myDef}Generalized evidence distance (GED)\end{myDef}
Assume there exists two GBPA, $m_{1}$ and $m_{2}$, on the frame of discernment, $A$. And the generalized evidence distance between $m_{1}$ and $m_{2}$ can be defined as:
\begin{equation}
	d_{GBPA}(m_{1},m_{2}) = \sqrt{\frac{1}{2}(\vec{m_{1}} - \vec{m_{2}})^{T}\overline{D}(\vec{m_{1}} - \vec{m_{2}})}
\end{equation}

in which the  $\overline{D}$ is a $2^{N} \times 2^{N}$ matrix whose elements are expressed as:
\begin{equation}
	\overline{D}(Q,W) = \frac{|Q \cap W|}{|Q \cup W|}
\end{equation}

\subsection{The Markov chain medel}
\begin{myDef}Markov property\end{myDef}Assume there exists a random series of conditions, $X_{n}:n>0$, namely the Markov chain, which is defined a space of probability $(\Omega, \Theta, P)$ in which $P$ denotes probability measure on the mapping between probability and conditions of events. Besides, all the probability is range from $0$ to $1$ in the restriction of the mapping. For any given moment $n$, any states $i,j \in S$ belongs to arbitrary space and any underlying and possible series of states $i_{0},i_{1},i_{2},...,i_{n-1}$ before the moment $n$, the Markov property can be defined as:
\begin{equation}
	\begin{split}
		P(X_{n+1} = j|X_{n} = i_{n},X_{n-1} = i_{n-1},...,X_{0} = i_{0}) =\\ P(X_{n+1} = j|X_{n} = i_{n})
	\end{split}
\end{equation}
\begin{myDef}The transition probability matrix of Markov chain\end{myDef}
Assume the transition probability is represented by $p_{ij}$ which means the possibility of a state $i$ transits to state $j$ from moment m to moment n. And it can be expressed as:
\begin{equation}
	P_{ij}(m,n) = P\{X_{n} = j|X_{m} = i\} = p_{ij}^{[m,n]}
\end{equation}

In order to simplify the procedure of presenting the probability $p_{ij}^{[m,n]}$, when the factor of time is not related with aspects discussed about the events, the icon of corresponding probability can be expressed as $p_{ij}$. And for a complete Markov chain, the transition probability matrix (TPM) can be defined as:
\begin{equation}
	TPM = \begin{bmatrix}
		p_{11} & p_{12} & p_{13} & ... & p_{1m} \\ 
		p_{21} & p_{22} & p_{23} & ... & p_{2m} \\ 
		... & ... & ... & ... & ... \\ 
		p_{m1} & p_{m2} & p_{m3} & ... & p_{mm} 
	\end{bmatrix}
\end{equation}

For every element contained in the matrix, some properties which are supposed to be satisfied can be defined as:
\begin{equation}
	p_{ij} \geq 0
\end{equation}
\begin{equation}
	\sum_{i = 1}^{m}\sum_{j = 1}^{m}p_{ij} = m
\end{equation}

\subsection{The method to obtain the transition probability matrix}

\begin{myDef}Cohort approach\end{myDef}
For any observed targets under given state $i$, the transition probability of transferring from state $i$ to state $j$ in a period of inspection can be defined as:
\begin{equation}
	p_{i,j,t} = \frac{N_{i,j,t}}{N_{i,t}}
\end{equation}

$N$ is the number of targets being observed and $t$ represents the moment when the observation terminates. However, for all the observed targets, it is necessary to weight all the transition probability in a period of observation and the process can be defined as:
\begin{equation}
	\widetilde{p}_{i,j} = \frac{\sum_{t=0}^{T}N_{i,t}p_{i,j,t}}{\sum_{t=0}^{T}N_{i,t}}
\end{equation}

In order to restrict the sum of element in every row exactly equal to $1$, a step of normalization is carried out and the detailed process is defined as:
\begin{equation}
	\overline{p}_{i,j} = \frac{	\widetilde{p}_{i,j}}{\sum_{j = 1}^{j = m}	\widetilde{p}_{i,j}}
\end{equation}

\subsection{Some entropy theories}
\begin{myDef}Deng entropy \cite{Deng2020ScienceChina}
\end{myDef}

Given a FOD, then the Deng entropy can be defined as:
\begin{equation}
	DE(r) = - \sum_{P \in 2^{A}}r(P)log_{2}\frac{r(P)}{2^{|P|-1}}
\end{equation}

In the expression of Deng entropy, $r$ is a mass function defined on the FOD and $P$ is set as a focal element. The mass of $r(P)$ indicates the support of belief of proposition $P$. Besides, the cardinality of proposition $P$ is represented by $|P|$.

\begin{myDef}Shannon entropy \cite{Shannon1948}
\end{myDef}

Given a series of distribution of probability, then the Shannon can be defined as:
\begin{equation}
	SH = -\sum_{n =1}^{m}p_{n}ln_{p_{n}}
\end{equation}

The sum of $p_{n}$ is equal to $1$.

\subsection{A kind of measure of similarity of evidences}
\begin{myDef}
	Deng et al.'s method \cite{DBLP:journals/dss/YongWZQ04}
\end{myDef}

Assume there exists two pieces of evidences $E_{i}$ and $E_{j}$ and the distances $d(E_{i},E_{j})$ can be calculated by the algorithm proposed in \cite{DBLP:journals/inffus/JousselmeGB01}. So, for any two pieces of evidences $E_{i}$ and $E_{j}$, the similarity between them can be calculated as:
\begin{equation}
	SIM(E_{n},E_{m}) = 1 - d(E_{n},E_{m})
\end{equation}

Then, for the whole body of evidences, a matrix manifest the similarity among evidences (SMM) can be given as:
\begin{equation}
	SMM = \begin{bmatrix}
		1&S_{12}&...&S_{1n}&...&S_{1m}\\
		\vdots&\vdots&...&\vdots&...&\vdots\\
		1&S_{f2}&...&S_{fn}&...&S_{fm}\\
		\vdots&\vdots&...&\vdots&...&\vdots\\
		1&S_{h2}&...&S_{hn}&...&1\\
	\end{bmatrix}
\end{equation}

Therefore, after obtaining the SMM, the corresponding support degree of pieces of evidences are defined as:
\begin{equation}
	SUP(E_{n}) = \sum_{m = 1}^{k} SIM(E_{n},E_{m})
\end{equation}

Then, the credibility degree CRD of according evidences $E_{L}$ are defined as:
\begin{equation}
	CRD_{L} = \frac{SUP_{E_{L}}}{\sum_{o = 1}^{k}SUP_{E_{o}}}
\end{equation}

which indicates an underlying relationship between evidences. To obtain a modified value of propositions of evidences, a parameter $MAE$ is defined as:
\begin{equation}
	MAE(r) = \sum_{u = 1}^{k}(CRD_{u} \times r_{u})
\end{equation}

After getting the modified values of propositions, if there exists $n$ pieces of evidence, then combine the evidences for $n-1$ times.

\subsection{Details of Z-numbers}
\begin{myDef}Z-numbers \cite{zadeh2011note} \end{myDef}A Z-number is composed of two fuzzy numbers to given a corresponding figure of the practical situations. Suppose there exists a kind of incident, $I$, and a pair of Z-number is relative with it. Then, the first dimension of Z-number is a probability measure of the incident, $I$, to happen and the second dimension is a check on the reliability of the judgement given in the first dimension. Therefore, the mathematics form of Z-numbers can be defined as:
\begin{equation}
	Z = (A,B)
\end{equation} 

\subsection{Sigmoid function}
Sigmoid function is often utilized as a activation function in neural network to map the variable into the range (0,1) and is defined as:
\begin{equation}
	S(x) = \frac{1}{1+e^{-x}}
\end{equation}

\section{Proposed Markov model for GET}
\subsection{The matrix form of GET (MFGET)}
Assume $L$ pieces of evidence is given on an FOD and the FOD is given as $A = \{a,b,c,d\}$. Then, the matrix of GBPAs distribution of evidences can be defined as:
\begin{equation}
	MFGET = \begin{bmatrix}
		P_{a}^{1} & P_{b}^{1} & P_{c}^{1} & P_{d}^{1} & ... & A_{2^{A}}^{1} \\ 
		P_{a}^{2} & P_{b}^{2} & P_{c}^{2} & P_{d}^{2} & ... & A_{2^{A}}^{2} \\ 
		... & ... & ... & ... & ...& ... \\ 
		P_{a}^{h} & P_{b}^{h} & P_{c}^{h} & P_{d}^{h} & ... & A_{2^{A}}^{h} \\ 
		P_{a}^{h+1} & P_{b}^{h+1} & P_{c}^{h+1} & P_{d}^{h+1} & ... & A_{2^{A}}^{h+1} \\ 
		... & ... & ... & ... & ...& ... \\ 
		P_{a}^{L} & P_{b}^{L} & P_{c}^{L} & P_{d}^{L} & ... & A_{2^{A}}^{L} \\ 
	\end{bmatrix}
\end{equation} 

For each column of the matrix, it can be regarded as a detailed figure about the situations of a specific proposition $P$. And the vectorial form of the $P$ with respect to any proposition $R, \ \ R \in 2^{A}$ can be defined as:
\begin{equation}
	VF = \{P_{R}^{1}, P_{R}^{2}, P_{R}^{3},..., P_{R}^{L}\}
\end{equation}

\begin{figure*}[h]
	\centering
	\includegraphics[scale=0.12]{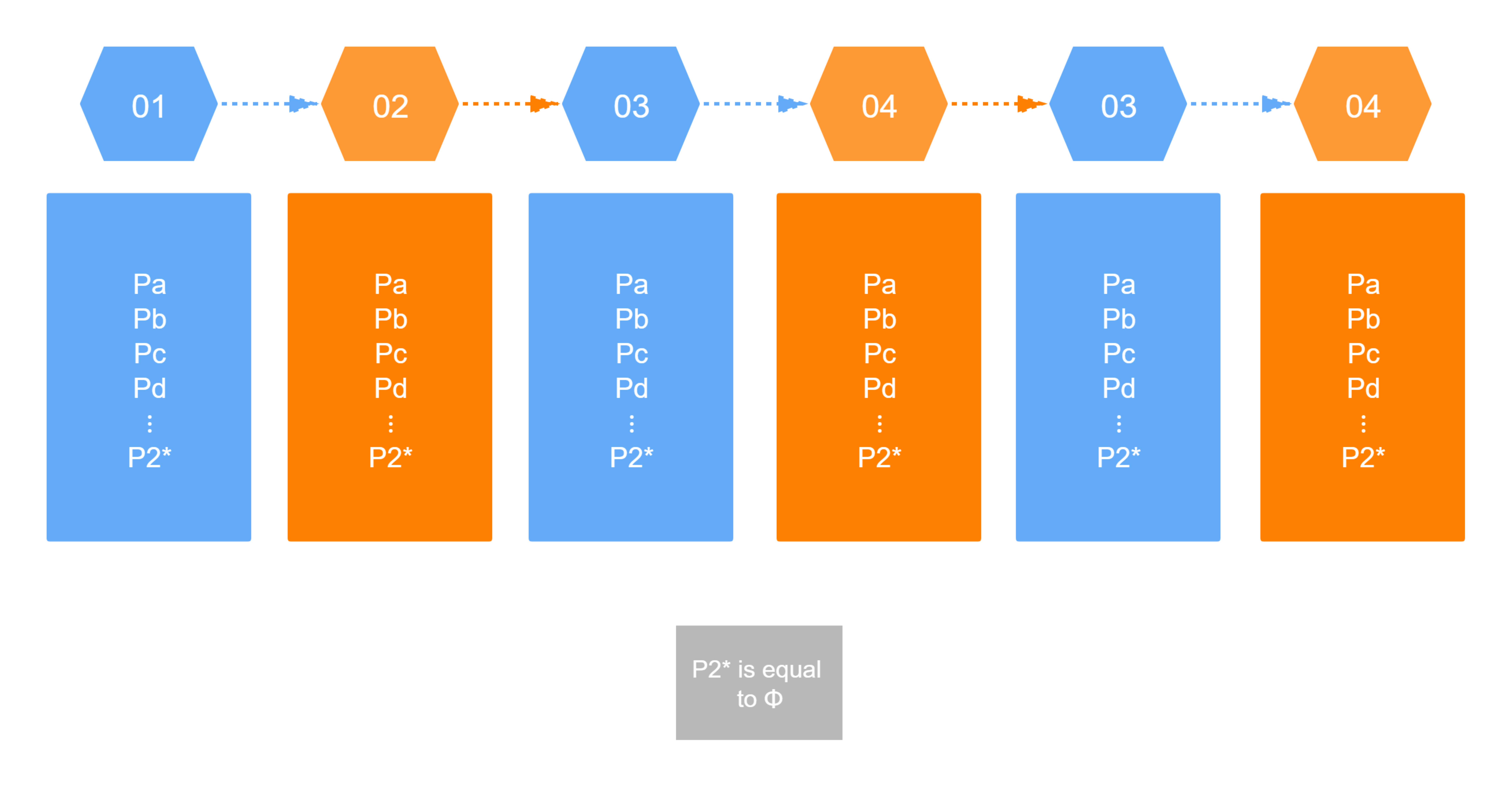}
	\caption{A matrix corresponding distribution for GET}
	\label{Fig4}
\end{figure*}

\subsection{An elimination designed to erase dirty data }
\begin{figure*}[h]
	\centering
	\includegraphics[scale=0.25]{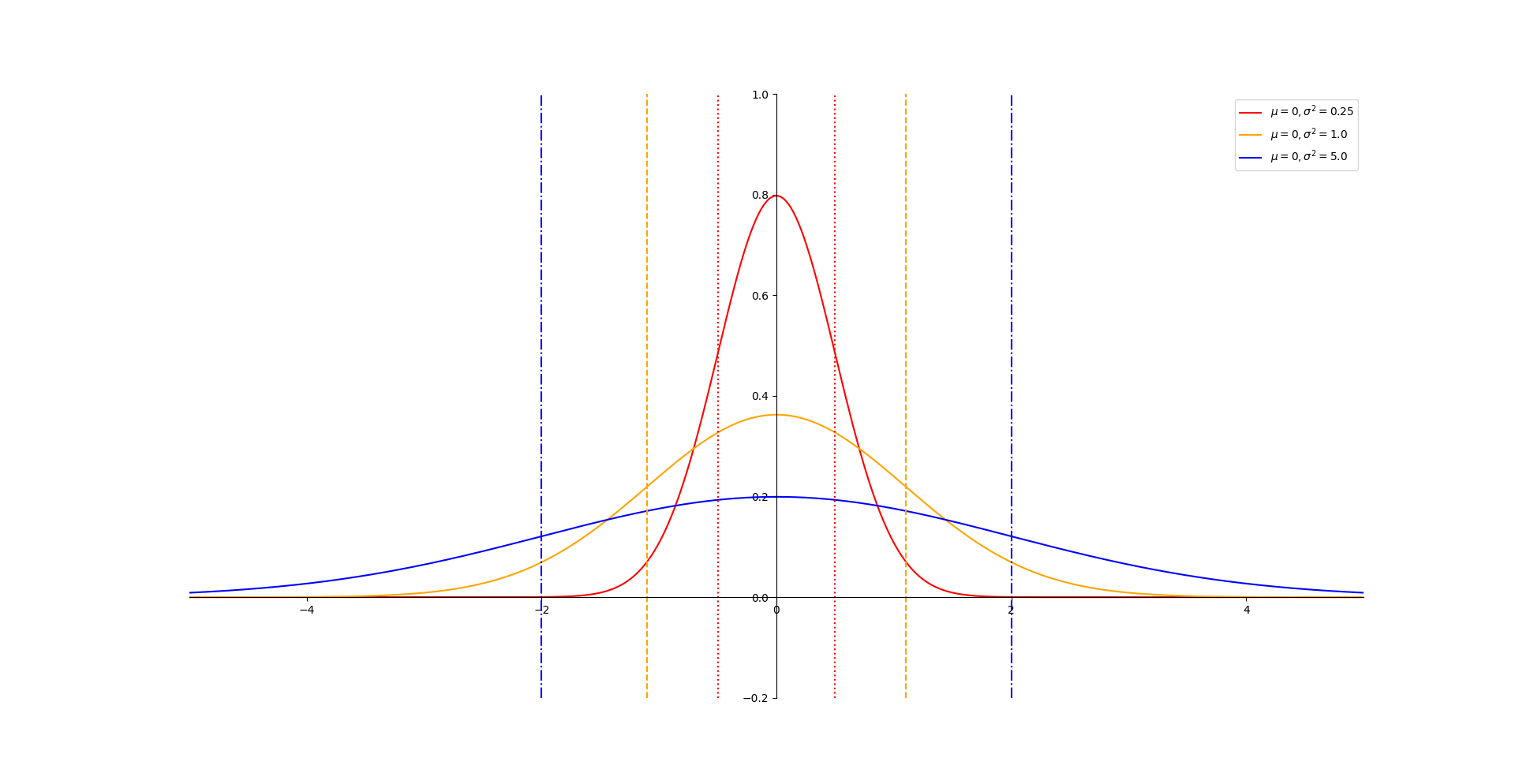}
	\caption{A simple view of the normal distribution}
	\label{Fig1}
\end{figure*}

An important probability distribution in the area of mathematics, physics and engineer is utilized in this customised elimination which is called as normal distribution. And the formula of it is defined as:
\begin{equation}
	f(x) = \frac{1}{\sqrt{2\pi}\sigma}exp(-\frac{(x-\mu)^{2}}{2\sigma^{2}})
\end{equation}

which can be also written as $X \sim N(\mu,\sigma^{2})$. Besides, the parameter $\mu$ and $\sigma$ can be defined as respectively:
\begin{equation}
	\mu = \sum_{k = 1}^{\infty} x_{k}p_{k} 
\end{equation}  
\begin{equation}
	\sigma^{2} = \frac{\sum(x_{k} - \mu)^{2}}{N}
\end{equation}

where the total variance is denoted by $\sigma^{2}$, $x_{k}$ represents each of variable, the mean of the whole is represented by $\mu$ and the number of the examples is $N$. The normal distribution is presented in Figure \ref{Fig1}.

Therefore, for a vectorial form of $P$, the corresponding parameters can be given as:
\begin{equation}
	\mu_{VF} = \frac{\sum_{i = 1}^{L} P_{R}^{i}}{L}
\end{equation}
\begin{equation}
	\sigma^{2}_{VF} = \frac{\sum(P_{R}^{i} - \mu_{VF})^{2}}{L}
\end{equation}

After both of the parameters are obtained, the evidences whose values of propositions provided lies in the corresponding range $(\mu_{VF} - \sigma_{VF}, \mu_{VF} + \sigma_{VF})$ are retained to ensure the data is effective and not conflicting with main body of evidences.

\subsection{A match between Markov chain (MC) and GET}
It can be obtained that a sequence of conditions is contained in the Markov chain which is very similar to the propositions, $P_{u},\ u \in 2^{A}$, contained in a frame of discernment (FOD). Let $IN$ denotes an incident which corresponding to a proposition defined on the frame of discernment which can be expressed as:
\begin{equation}
	IN_{n} = P_{u},\ \ \ \ u \in 2^{A}, n \ \ in\ \ (1,|2^{A}|)
\end{equation}

Suppose there exists a series of incidents and the chain of incidents can be given as:
\begin{equation}
	MC = \{IN_{1}, IN_{2},IN_{3},...,IN_{|2^{A}|}\}, \ \ \ \ IN_{|2^{A}|} = \emptyset
\end{equation}

According to the definition of the GET, each proposition contained in the FOD is allocated a mass of GBPA which means the proposition owns a possibility to take place. In the most optimistic situation, all of the proposition can happen which indicates that when one incident occur then it can be transferred into another condition that the same or a different proposition to take place. It can be appropriately and accurately described by a transiting chain matrix (TCM) which can be defined as:
\begin{small}
\begin{equation}
\rotatebox{90}{$TCM = \begin{bmatrix}
		IN_{1} \longrightarrow IN_{1} &IN_{1} \longrightarrow IN_{2}  & ... & IN_{1} \longrightarrow IN_{2^{A}} \\ 
		IN_{2} \longrightarrow IN_{1} &IN_{2} \longrightarrow IN_{2}  & ... & IN_{2} \longrightarrow IN_{2^{A}} \\ 
		... & ... & ... & ...   \\ 
		IN_{F} \longrightarrow IN_{1} &IN_{F} \longrightarrow IN_{2}  & ... & IN_{F} \longrightarrow IN_{2^{A}} \\ 
		IN_{F+1} \longrightarrow IN_{1} &IN_{F+1} \longrightarrow IN_{2}  & ... & IN_{F+1} \longrightarrow IN_{2^{A}} \\ 
		... & ... & ... & ...   \\ 
		IN_{2^{A}} \longrightarrow IN_{1} &IN_{2^{A}} \longrightarrow IN_{2}  & ... & IN_{2^{A}} \longrightarrow IN_{2^{A}} \\ 
	\end{bmatrix}$}
\end{equation}
\end{small}
where $	IN_{i} \longrightarrow IN_{j}$ represents the probability of the transformation from incident $i$ to incident $j$. In order to simplify the process of manifesting each details of propositions' changes, a icon $T_{ij}$ is proposed and the matrix $TCM$ can be rewritten as:
\begin{equation}
\rotatebox{90}{$TCM = \begin{bmatrix}
		T_{11} &T_{12}  & ...&T_{1n}&T_{1(n+1)}&... & T_{1|2^{A}|} \\ 
		T_{21} &T_{22}  & ...&T_{2n}&T_{2(n+1)}&... & T_{2|2^{A}|} \\ 
		... & ... & ... & ... & ... & ... & ...  \\ 
		T_{k1} &T_{k2}  & ...&T_{kn}&T_{k(n+1)}&... & T_{k|2^{A}|} \\ 
		T_{(k+1)1} &T_{(k+1)2} & ...&T_{(k+1)n}&T_{(k+1)(n+1)}&... & T_{(k+1)|2^{A}|} \\ 
		... & ...  & ...& ... & ... & ... & ...  \\ 
		T_{|2^{A}|1} &T_{|2^{A}|2}  & ...&T_{|2^{A}|n}&T_{|2^{A}|(n+1)}&... & T_{|2^{A}||2^{A}|} \\ 
	\end{bmatrix}$}
\end{equation}

Besides, one restriction is supposed to be satisfied which is defined as:
\begin{equation}
	\sum_{b = 1}^{b = |2^{A}|}T_{ib} = 1 
\end{equation}

\begin{figure*}[h]
	\centering
	\includegraphics[scale=0.38]{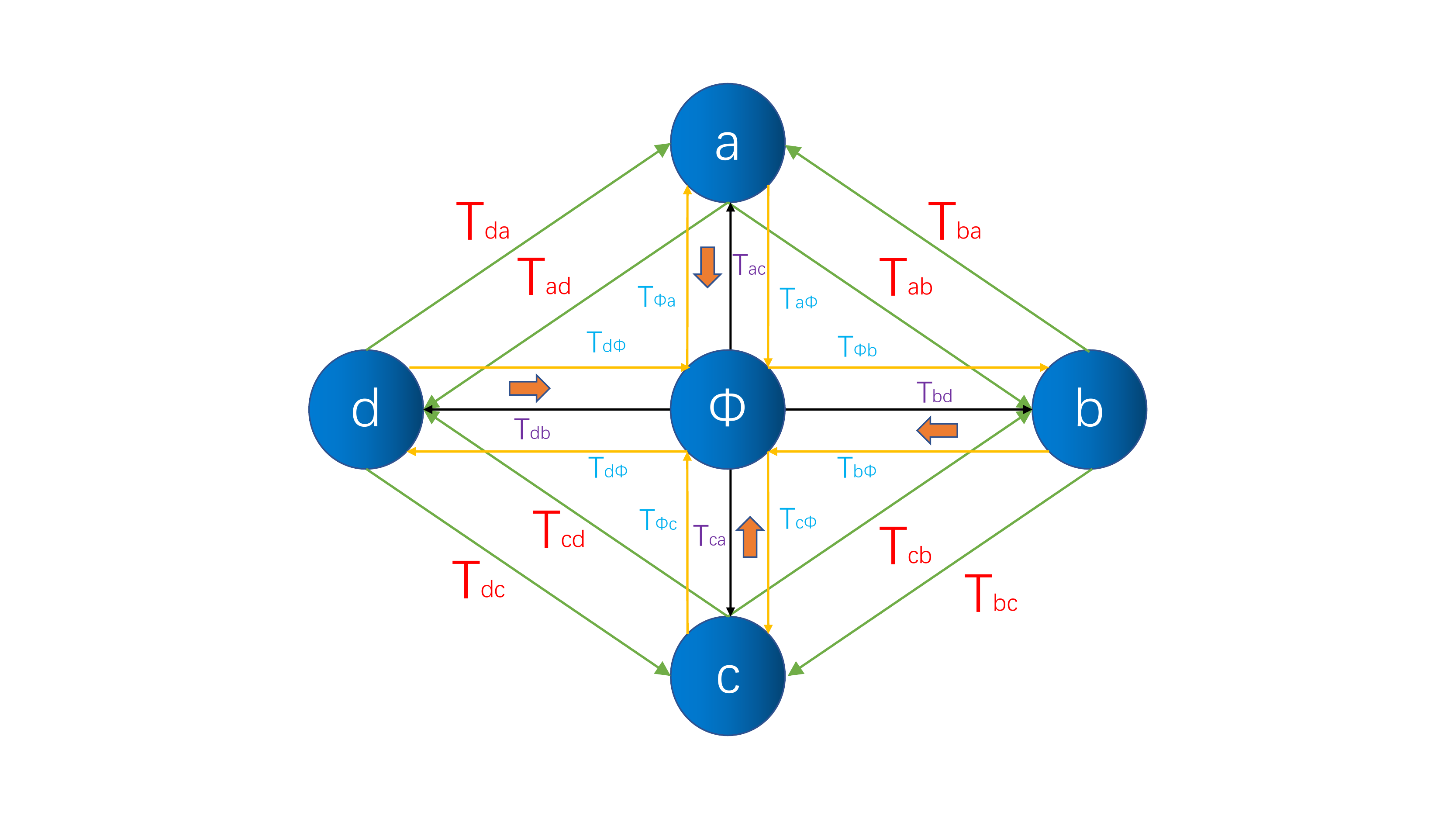}
	\caption{A Markov process of a FOD}
	\label{Fig2}
\end{figure*}

Assume there exists a FOD which is defined as $\Theta = \{a,b,c,d,\emptyset\}$, the Markov process of the FOD is given in Figure \ref{Fig2}.

\subsection{The definition of transition probability}

A series of GBPAs are given in a piece of evidence on the definition of FOD and it can be concluded that the higher the value of GBPA of a proposition is, the bigger probability the incident corresponding to proposition is, which also affect the next state of development of things. Therefore, it is reasonable and rational to regard that if the absolute mass of subtraction of values of GBPAs of two propositions is high, then the underlying possibility of a transferring between the two proposition is low. Then, a cost function to measure the cost in the process of transferring can be defined as:
\begin{equation}
	Cost = \frac{1}{1+ e^{5(m(a) - m(b))}} \ \ \ \ \ \ a \longrightarrow b
\end{equation}

\begin{figure}[h]
	\centering
	\includegraphics[scale=0.45]{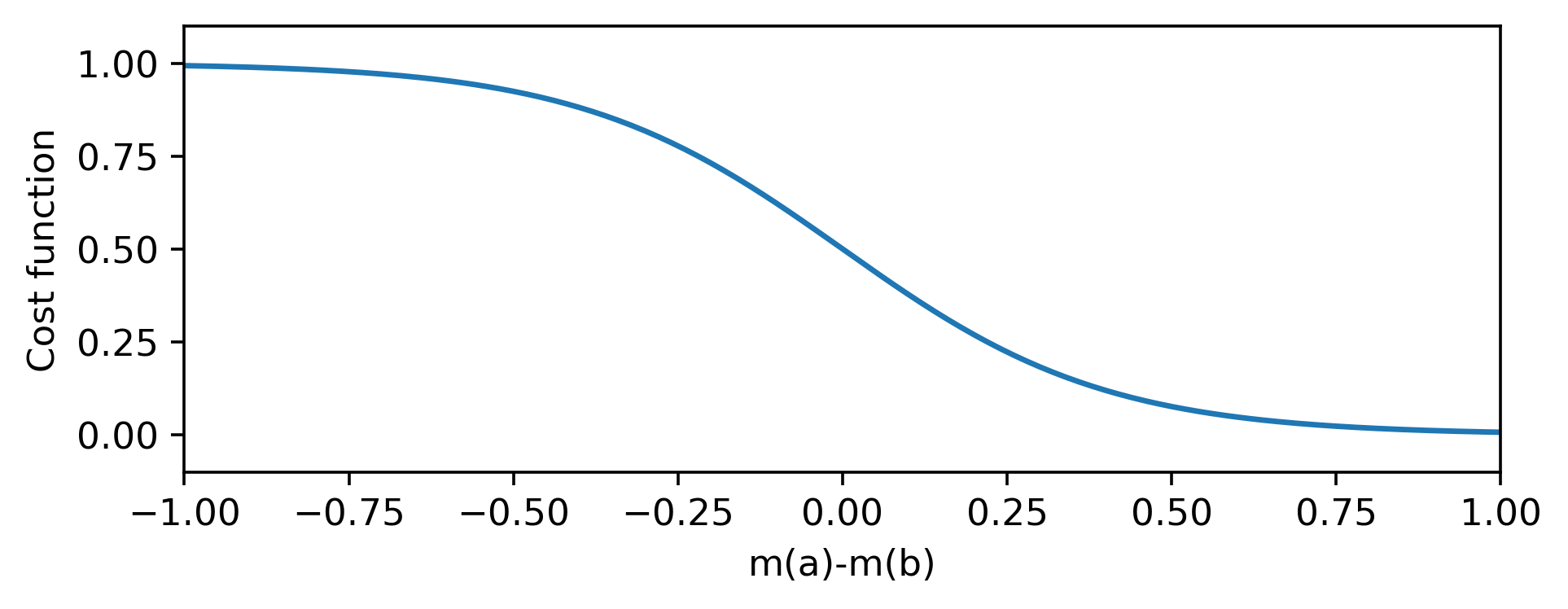}
	\caption{A transformation for MC using modified sigmoid function}
	\label{Fig5}
\end{figure}

And the settings for the cost function is based on the theory of increase of entropy. It can be easily concluded that everything is becoming more and more uncertainty simultaneously which is corresponding to the phenomenon that the state with a high mass transfers to a state which owns a low mass. Therefore, the cost of the process is regarded as relatively low. On the other side, if one uncertainty state with a very low mass intends to become a certain state with a relatively high mass, the cost is reasonably much bigger due to the property that everything is going to be more and more chaotic.

So, with respect to the range of the mass of propositions given in a piece of evidence, the transition probability can be defined as:
\begin{equation}
	TP = \frac{Cost_{i}}{\sum_{k = 1}^{|2^{A}|}Cost_{k}}
\end{equation}

\subsection{Concomitant sets for incidents in MC}
The transition probabilities of a incident can be obtained through the procedure proposed above. Then, a concomitant sets can be given to describe the condition of an incident, which can be defined as:
\begin{equation}
	CS = \{IN_{i} | TP_{ij},...,TP_{ik}, ..., TP_{i|2^{A}|}\}
\end{equation}

It can be easily concluded that if the sum of transition probability of the $IN_{i}$ is relatively low, then the certainty of the proposition is relatively high which means the next state of the sequence of things is probably still the same incident. Therefore, the certainty measure of the $IN_{i}$ can be defined as:
\begin{equation}
	CM = \sum_{j \neq i}\frac{1}{TP_{ij} \rightarrow Cost}
\end{equation}

\begin{figure*}[h]
	\centering
	\includegraphics[scale=0.38]{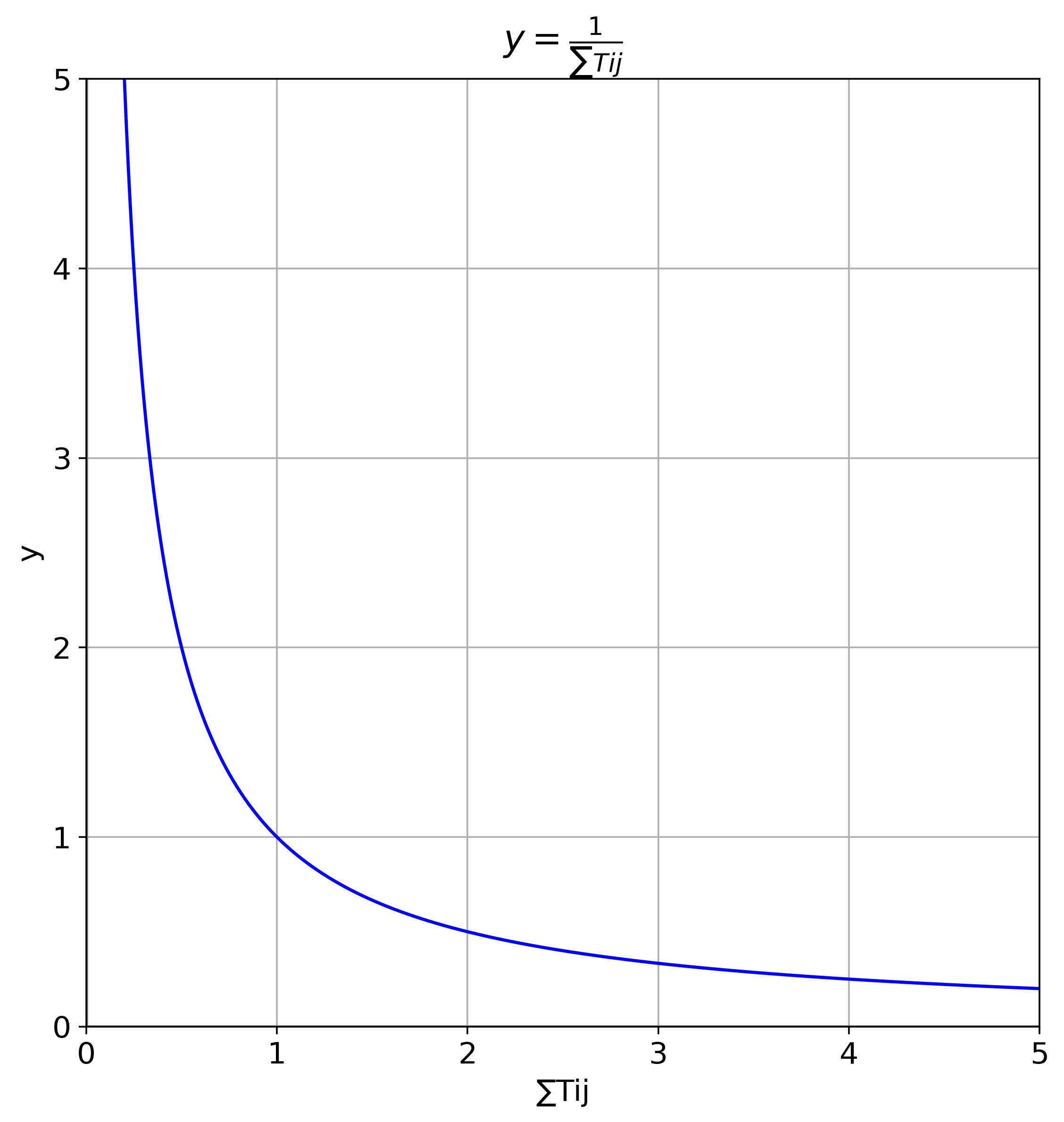}
	\caption{The process of fluctuation of one element of CM}
	\label{Fig3}
\end{figure*}

In the expression of the CM, the bigger the value of CM is, the more certain the proposition or incident is which can be regarded as a kind of steady state of the evidence given. If a evidence owns a prominent CM of one proposition, then the information transferred by the evidence is explicit and not confusing.

\subsection{Belief function for MMGET (BF)}

Assume there exists a $IN_{i}$ and its CS is given as $\{IN_{i} | TP_{ij},...,TP_{it}, ..., TP_{i|2^{A}|}\}$. According to the definition of GBF, the specially designed belief function for MMGET can be defined as:
\begin{equation}
	\begin{split}
			BF = \{\sum_{IN_{k} \subseteq IN_{i}}IN_{k}|\frac{TP_{ij} + TP_{kj} + ...}{n_{1}},...,\\ \frac{TP_{it} + TP_{kt} + ...}{n_{2}},...,\frac{TP_{i|2^{A}|} + TP_{k|2^{A}|}+...}{n_{3}}\}
	\end{split}
\end{equation}

The variable $n_{i}$ is the number of the process of the transferring from one state to another state. 

A simple example is designed to illustrate the process of obtaining the value of BF. Assume there exists three CSs which is given as $\{A = 0.1|0.4,0.3,0.3\}$, $\{\{A,B\} = 0.3|0.35,0.2,0.4\}$, $\{\{A,B,C\} = 0.2|0.2,0.45,0.35\}$. Then, with respect to the incident $\{A,B,C\}$, the BF of it can be calculated as:

$BF_{\{A,B,C\}} = \{0.2 + 0.3 + 0.1| \frac{0.4+0.35+0.2}{3}, \frac{0.3+0.2+0.45}{3},\frac{0.3+0.4+0.35}{3}\} = \{0.6|0.3166,0.3166,0.3500\}$

\subsection{Plausible function for MMGET (PF)}

Assume there exists a $IN_{i}$ and its CS is given as $\{IN_{i} | TP_{ij},...,TP_{it}, ..., TP_{i|2^{A}|}\}$. According to the definition of GPF, the specially designed belief function for MMGET can be defined as:
\begin{equation}
	\begin{split}
		PF = \{\sum_{IN_{k} \cap IN_{i} \neq \emptyset}IN_{k}|\frac{TP_{ij} + TP_{kj} + ...}{n_{1}},...,\\ \frac{TP_{it} + TP_{kt} + ...}{n_{2}},...,\frac{TP_{i|2^{A}|} + TP_{k|2^{A}|}+...}{n_{3}}\}
	\end{split}
\end{equation}

The variable $n_{i}$ is the number of the process of the transferring from one state to another state. 

A simple example is designed to illustrate the process of obtaining the value of BF. Assume there exists three CSs which is given as $\{A = 0.1|0.4,0.3,0.3\}$, $\{\{A,B\} = 0.3|0.35,0.2,0.4\}$, $\{\{B\} = 0.2|0.2,0.45,0.35\}$. Then, with respect to the incident $\{A\}$, the BF of it can be calculated as:

$PF_{\{A\}} = \{0.1 + 0.3| \frac{0.4+0.35}{2}, \frac{0.3+0.2}{2},\frac{0.3+0.4}{2}\} = \{0.4|0.375,0.25,0.35\}$

\subsection{Combination rule for MMGET}
In the MMGET, some according mass is allocated to a $IN_{i}$. A certainty measure is proposed in this paper called CM which can be utilized as a credibility proof of the specific proposition. For a piece of evidence after modification, $E$, the allocation of mass of propositions can be defined as:
\begin{equation}
	E = \{\frac{IN_{1}CM_{1}}{\sum_{i = 1}^{|2^{A}|}IN_{i}CM_{i}},\frac{IN_{2}CM_{2}}{\sum_{i = 1}^{|2^{A}|}IN_{i}CM_{i}},...,\frac{IN_{|2^{A}|}CM_{|2^{A}|}}{\sum_{i = 1}^{|2^{A}|}IN_{i}CM_{i}}\}
\end{equation}

In the process of combination, the element $\emptyset$ is treated differently due to its effect in indicating the extend of completion of the whole OFD. The detailed process of combination for MMGET can be defined as:
\begin{equation}
	K = \sum_{IN_{o}\cap IN_{p} = \emptyset}\frac{IN_{o}CM_{o}}{\sum_{i = 1}^{|2^{A}|}IN_{i}CM_{i}}\frac{IN_{p}CM_{p}}{\sum_{g = 1}^{|2^{A}|}IN_{g}CM_{g}}
\end{equation}
\begin{equation}
	IN_{|2^{A}|} = \frac{IN_{|2^{A}|}CM_{|2^{A}|}}{\sum_{i = 1}^{|2^{A}|}IN_{i}CM_{i}}\frac{IN_{|2^{A}|}CM_{|2^{A}|}}{\sum_{g = 1}^{|2^{A}|}IN_{g}CM_{g}} = m(\emptyset)
\end{equation}
\begin{equation}
	IN_{k} = \frac{(1 - m(\emptyset))\sum_{IN_{o}\cap IN_{p} = IN_{k}}\frac{IN_{o}CM_{o}}{\sum_{i = 1}^{|2^{A}|}IN_{i}CM_{i}}\frac{IN_{p}CM_{p}}{\sum_{g = 1}^{|2^{A}|}IN_{g}CM_{g}}}{1 - K}
\end{equation}

\begin{equation}
	IN_{|2^{A}|} = 1\ \  if \ \ and \ \ only \ \ if \ \ K = 1
\end{equation}

A simple example is designed to illustrate the process of the combination. Assume there exists two series of distribution of $E$ which is given as $E_{1} = \{A_{1} = 0.3, B_{1} = 0.6, \emptyset_{1} = 0.1\}$ and $E_{2} = \{A_{2} = 0.6, B_{2} = 0.3, \emptyset_{2} = 0.1\}$. And the process of combination can be given as:

$K = A_{1}*(B_{2} + \emptyset_{2}) + B_{1}*(A_{2}+\emptyset_{2}) +  \emptyset_{1} *(A_{2}+B_{2}+\emptyset_{2}) = 0.64$ 

$m(\emptyset) = 0.1 * 0.1 = 0.01$

$IN_{A} = \frac{(1-0.01)(0.3 * 0.6)}{1 - 0.64} = 0.495$

$IN_{B} = \frac{(1-0.01)(0.6 * 0.3)}{1 - 0.64} = 0.495$

\subsection{Distances of elements in the MP}
\subsubsection{Distance of MCs in MMGET}
Assume there exists two MCs, $c_{1}$ and $c_{2}$, and the vectorial form of chains is given as $\{IN_{1},IN_{2},IN_{3},IN_{4}\}$. Then, the distance measure for MC is defined as:
\begin{equation}
	d_{MC}(c_{1},c_{2}) = \sqrt{\frac{1}{2}(\vec{c_{1}} - \vec{c_{2}})^{T}\widehat{D}(\vec{c_{1}} - \vec{c_{2}})}
\end{equation}

in which the  $\widehat{D}$ is a $2^{N} \times 2^{N}$ matrix whose elements are expressed as:
\begin{equation}
	\widehat{D} = \frac{2^{|IN_{i} \cap IN_{j}|} - 1}{\sqrt{(2^{|IN_{i}|} - 1)(2^{|IN_{j}|} - 1)}}, \ \ \ \ i,j \ \ in \ \ (1,|2^{A}|)
\end{equation}

A simple example is revised to illustrate the effectiveness of the proposed method of measuring the distance of MC. Assume there exists two pieces of MCs which is given as $c_{1} = \{A_{1} = 0.3, B_{1} = 0.4, C_{1} = 0.1, \emptyset = 0.2\}$ and  $c_{2} = \{A_{2} = 0.1, B_{2} = 0.3, C_{2} = 0.5, \emptyset = 0.1\}$. Then, the process of calculating the distance can be given as:

$d_{MC}(c_{1},c_{2}) =\\ \sqrt{\frac{1}{2}(0.2,0.1,-0.4,0.1)^{T}\begin{pmatrix} 
		
		1 & 0 &0 &0 \\
		
		0 & 1&0&0\\
		0&0&1&0\\
		0&0&0&1
		
	\end{pmatrix}(0.2,0.1,-0.4,0.1)}\\ = 0.3316$

\subsubsection{Distance of CSs in MMGET}
Assume there exists two CSs, $a_{1}$ and $a_{2}$, and the vectorial form of chains is given as $\{CS_{IN_{1}},CS_{IN_{2}},CS_{IN_{3}},CS_{IN_{4}}\}$. Then, the distance measure for CS is defined as:
\begin{equation}
	d_{MC}(a_{1}^{MC},a_{2}^{MC}) = \sqrt{\frac{1}{2}(\vec{a_{1}^{MC}} - \vec{a_{2}^{MC}})^{T}D(\vec{a_{1}^{MC}} - \vec{a_{2}^{MC}})}
\end{equation}
Besides, for the corresponding series of TP of propositions, the distance between TPs is defined as:

\begin{equation}
	d_{TP}(a_{1}^{TP},a_{2}^{TP}) = \sqrt{\frac{1}{2}(\vec{a_{1}^{TP}} - \vec{a_{2}^{TP}})^{T}D(\vec{a_{1}^{TP}} - \vec{a_{2}^{TP}})}
\end{equation}

\textbf{Note: }For the $d_{TP}$, the incidents of the matrix $D$ is the ones corresponding to the transferred state.

in which the  $D$ is a $2^{N} \times 2^{N}$ matrix whose elements are expressed as:
\begin{equation}
	D = \frac{2^{|IN_{i} \cap IN_{j}|} - 1}{\sqrt{(2^{|IN_{i}|} - 1)(2^{|IN_{j}|} - 1)}}, \ \ \ \ i,j \ \ in \ \ (1,|2^{A}|)
\end{equation}

All in all, the distance of CSs is defined as:
\begin{equation}
	d_{CS} = d_{MC}(a_{1}^{MC},a_{2}^{MC})+ \sum d_{TP}(a_{1}^{TP},a_{2}^{TP})
\end{equation}

The process of calculation of distances is the same to the procedure provided in the section of distances of MCs.

\subsubsection{Distance of modified evidences in MMGET}
Assume there exists two pieces of modified evidences, $e_{1}$ and $e_{2}$, whose vectorial form are given as $\{\frac{IN_{1}CV_{1}}{\sum_{i = 1}^{|2^{A}|}IN_{i}CV_{i}},\frac{IN_{2}CV_{2}}{\sum_{i = 1}^{|2^{A}|}IN_{i}CV_{i}},...,\frac{IN_{|2^{A}|}CV_{|2^{A}|}}{\sum_{i = 1}^{|2^{A}|}IN_{i}CV_{i}}\}$. Then, the measure of distance of the modified evidences is defined as:
\begin{equation}
	d_{ME}(e_{1},e_{2}) = \sqrt{\frac{1}{2}(\vec{e_{1}} - \vec{e_{2}})^{T}\widetilde{D}(\vec{e_{1}} - \vec{e_{2}})}
\end{equation}

in which the  $\widetilde{D}$ is a $2^{N} \times 2^{N}$ matrix whose elements are expressed as:
\begin{equation}
	\widetilde{D} = \frac{2^{|IN_{i} \cap IN_{j}|} - 1}{\sqrt{(2^{|IN_{i}|} - 1)(2^{|IN_{j}|} - 1)}}, \ \ \ \ i,j \ \ in \ \ (1,|2^{A}|)
\end{equation}

The process of calculation of distances is the same to the procedure provided in the section of distances of MCs.

\subsection{Measure of similarity of original evidences}
It is necessary to calculate a degree of divergence of a piece of evidence among the main body of evidences. In this part, the difference of the probability assignment of an incident, the corresponding series of transition probability and modified evidences is taken into consideration. Therefore, the similarity measure of original evidences is defined as:
\begin{equation}
	SM = \frac{\sum (d_{ME}^{VF_{i}}+ \frac{1}{2}\sum d_{CS}^{VF_{i}})^{2}}{(d_{ME}^{VF_{j}}+ \frac{1}{2}\sum d_{CS}^{VF_{j}})^{2}} = \frac{\sum DI^{VF_{i}}}{DI^{VF_{j}}}
\end{equation}

From the formula defined above, it can be concluded that if the distance to other evidences of a piece of evidence is relatively big which means the one judged is much more different than other ones, then the value of parameter $SM$ is relatively lower in turn. A simple example is designed to illustrate the effectiveness of the $SM$.

Assume there exists a series of group of $(d_{ME}^{VF_{i}}+ \frac{1}{2}\sum d_{CS}^{VF_{i}})^{2}$ is given respectively as $\{dsum_{VF_{1}} = 0.3,dsum_{VF_{2}} =0.4,dsum_{VF_{3}} =0.5,dsum_{VF_{4}} =0.7\}$. Then, the degree of reliability of each VF can be calculated as:

$SM_{VF_{1}} = \frac{0.09+0.16+0.25+0.49}{0.09} = 11.0000$

$SM_{VF_{2}} = \frac{0.09+0.16+0.25+0.49}{0.09} = 6.1875$

$SM_{VF_{3}} = \frac{0.09+0.16+0.25+0.49}{0.09} = 3.9600$

$SM_{VF_{4}} = \frac{0.09+0.16+0.25+0.49}{0.09} = 2.0204$

\begin{figure*}[h]
	\centering
	\includegraphics[scale=0.3]{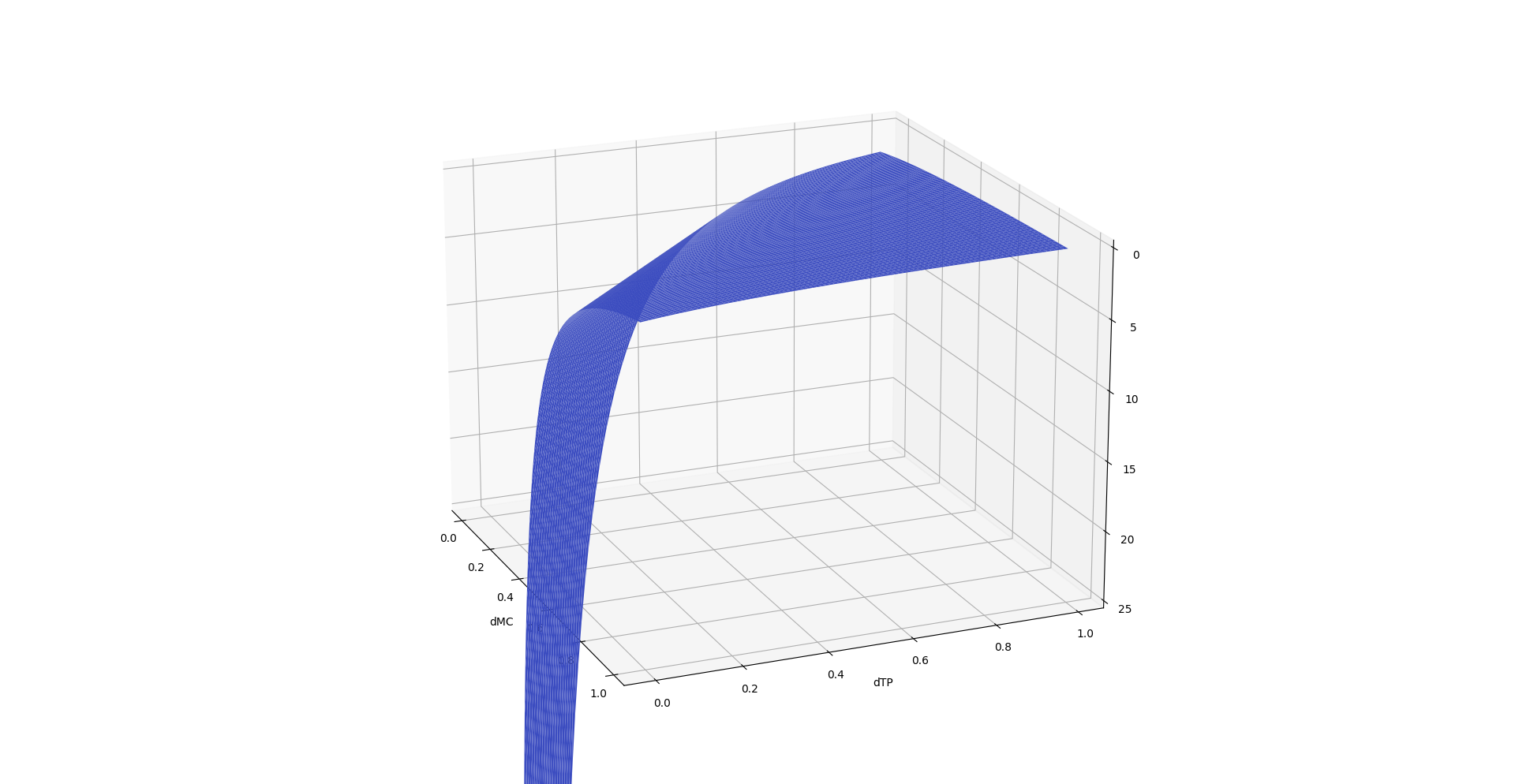}
	\caption{The fluctuation of SM}
	\label{Fig6}
\end{figure*}

\textbf{Example} Assume there exists a series of distribution of $d_{ME}^{VF_{i}}$ and $d_{CS}^{VF_{i}}$, three propositions contained in the frame of discernment and three piece of evidences obtained. And the process of fluctuation of the variable SM is given in Figure \ref{Fig6}. 

\subsection{Certainty measure of FOD}
\begin{figure*}[h]
	\centering
	\includegraphics[scale=0.3]{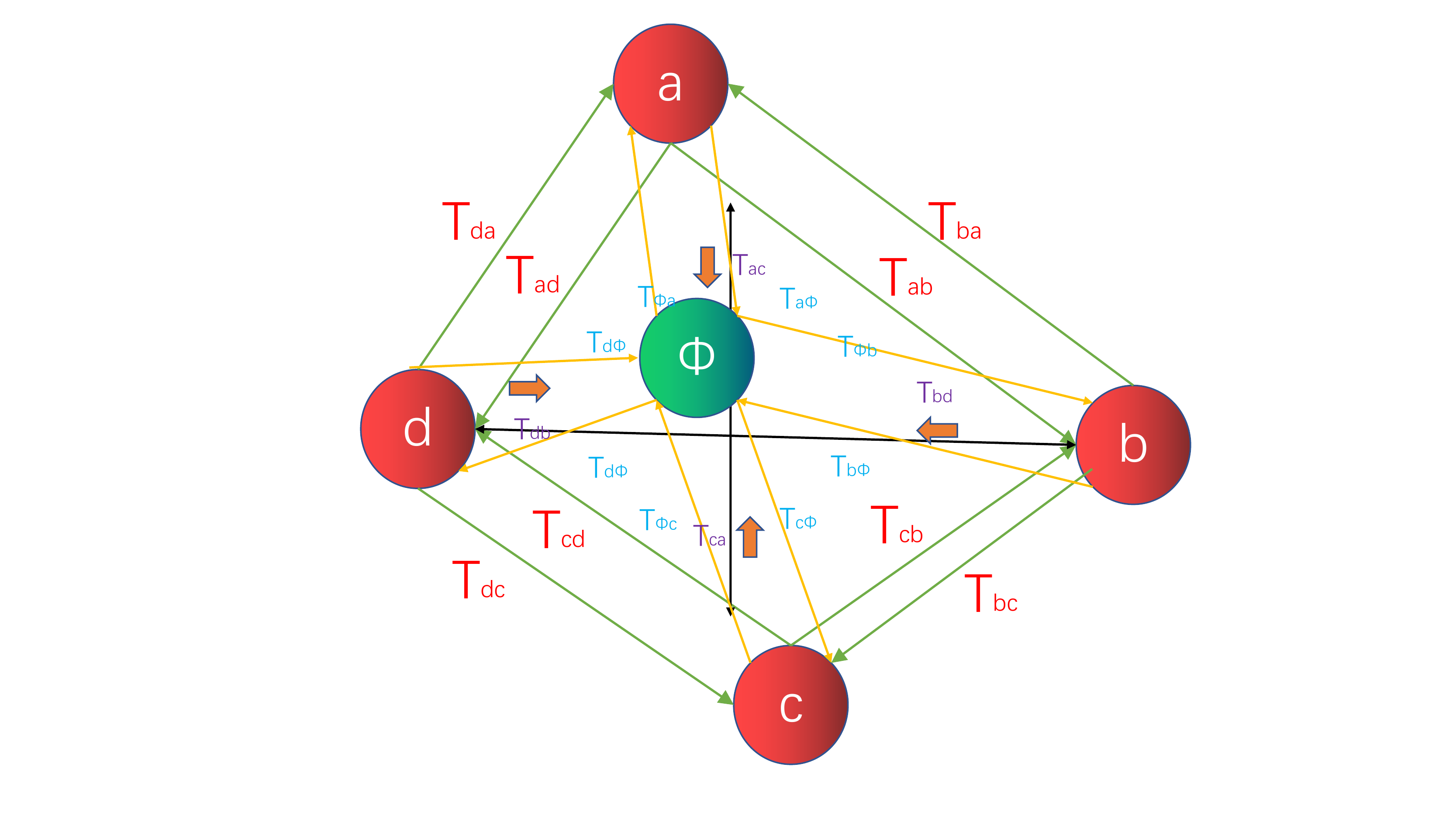}
	\caption{The fluctuation of SM}
	\label{Fig8}
\end{figure*}

In GET, the value of $m(\emptyset)$ indicate the degree of uncertainty of the frame of discernment provided. Therefore, an innovative method to recognize the uncertainty is designed to manifest this kind of phenomenon. And first, with the changes of values of propositions and empty set, the degree of uncertainty of the frame of discernment is also changed. A simple Figure \ref{Fig8} is drawn to present this kind of fluctuation.

It can be easily concluded that if all of the propositions except the empty set are much more easily to transfer from the original state to the state of empty set, then it tends to regard the FOD as incomplete, which also indicates that if the sum of transition probability is bigger, then the FOD is more incomplete. To reflect the level of uncertainty of the FOD, an uncertainty-recognition entropy is defined as:
\begin{equation}
	CRE = \sum_{i = 1}^{n}TP_{IN_{i} \rightarrow IN_{j}}^{2}*2^{(|IN_{i}|-|IN_{j}|)}
\end{equation}

\textbf{Note: }If the $\emptyset$ is the object a proposition transferred to, then the value of $|IN_{i}|$ is set as the number of biggest cardinality of propositions. And if $IN_{i} = \emptyset$, then the value of $|IN_{i}|$ is set as the number of smallest cardinality of propositions. The two settings are designed to help better present the degree of uncertainty of FOD.  

From the definition of the formula, if the transition probability gets bigger, then the CRE also gets bigger which indicates that if some states corresponding to some propositions are easily to transit to the state of empty set, then the FOD tends to be more unsteady. What is expected to be pointed out that, if a state of multiple propositions transfers, then the degree of uncertainty reduces which means a much bigger cost in the process of transition and shows that the transition to state of empty set is much more attractive which means there exists a more unsteady FOD. 

\begin{figure*}[h]
	\centering
	\includegraphics[scale=0.65]{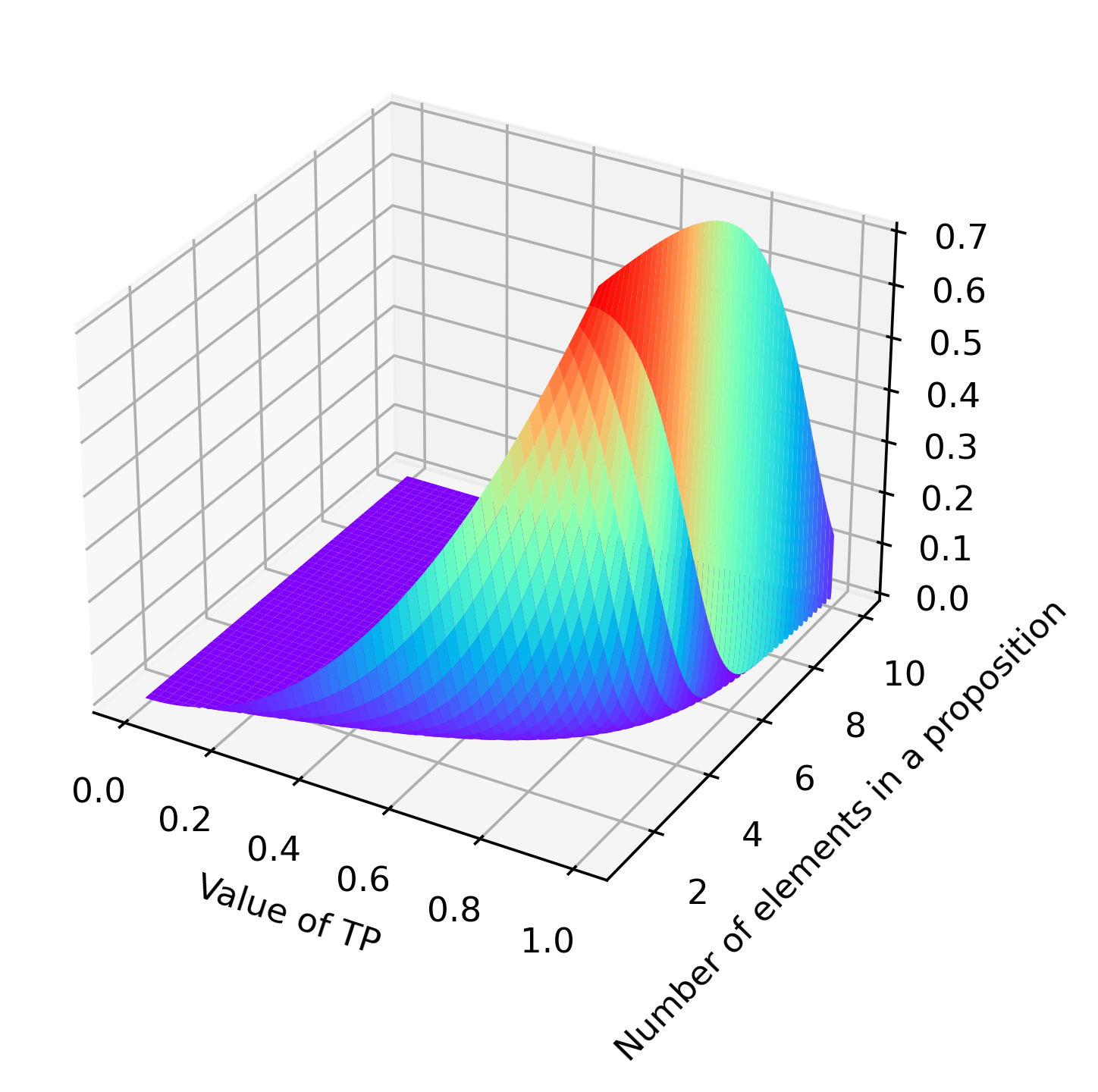}
	\caption{The fluctuation of CRE}
	\label{Fig9}
\end{figure*}

Assume the elements contained in a proposition is less than 10, and a simple Figure \ref{Fig9} is drawn to indicate the fluctuation of values of CRE.

Besides, a simple example is revised to illustrate the process of obtaining the CRE. Assume there exists a series of TPs whose according propositions are single proposition which is given as $\{0.2,0.4,0.5\}$. And the CRE can be obtained as:

$CRE = 0.2^{2} * 2^{1 -1} + 0.4^{2} * 1 + 0.5^{2} * 1 = 0.45$

\begin{figure*}[h]
	\centering
	\includegraphics[scale=0.09]{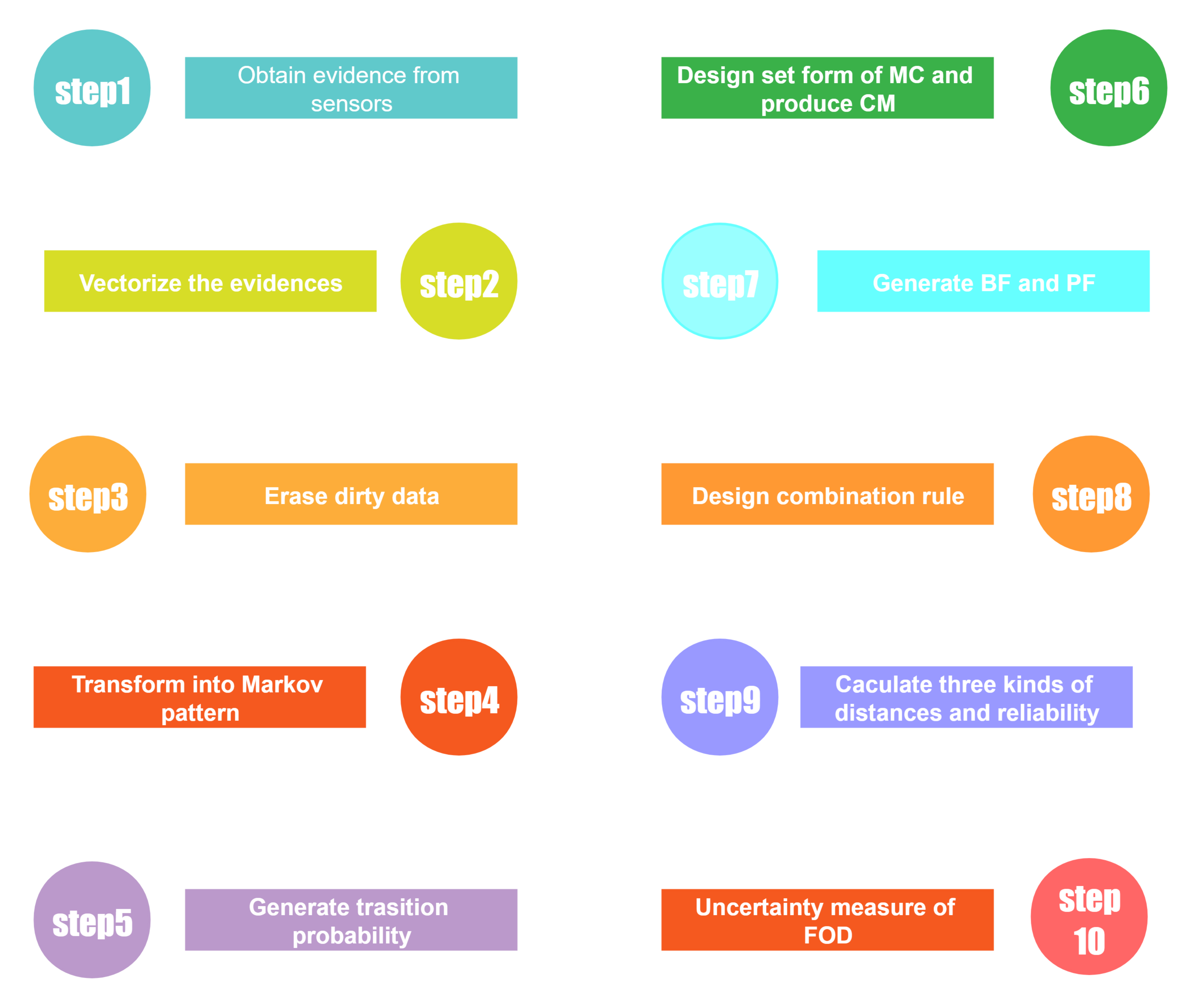}
	\caption{The detailed process of MF for GET}
	\label{Fig7}
\end{figure*}

In the last, a flow chart is designed to illustrate the details of proposed model. All of the process is given in Figure \ref{Fig7}.

\section{Numerical example}
\textbf{Example: }Illustration of the similarity measure

Assume there exists three pieces of evidences which owns 3 propositions whose FOD can be defined as $\Theta = \{a,b,c\}$. The values of propositions contained in the evidences are listed in Table \ref{table111}. Besides, the corresponding cost and TPs are listed in Table \ref{tableuu} and \ref{table33}. According to the definition of similarity measure, three kinds of distances are supposed to be calculated to obtain a base to calculate the value of similarity. Then, the three kinds of distances, $d_{MC}$, $d_{TP}$ and $d_{ME}$, are produced by the uniform method of measure of distance which are given in Table \ref{table334}, \ref{table335} and \ref{table336}. In the last, the final judgment on the similarity of evidences are listed in Table \ref{table337}.

\begin{table}[h]\footnotesize
	\centering
	\caption{The evidences given by sensors}
		\begin{tabular}{c c c c c c c c c c c c c c c c}\hline
			$Evidences$ & \multicolumn{3}{c}{$Values \ \ of \ \ propositions $} \\\hline
			$Evidence_{1}$& $\{a\}$ & $\{b\}$ &$\{c\}$ \\
			& $0.1$ & $0.6$&$0.3$ \\
			$Evidence_{2}$& $\{a\}$ & $\{b\}$ &$\{c\}$ \\
			& $0.2$ & $0.1$&$0.7$ \\
			$Evidence_{3}$& $\{a\}$ & $\{b\}$ &$\{c\}$\\
			& $0.5$ & $0.3$&$0.2$ \\
			\hline
		\end{tabular}
	
	\label{table111}
\end{table}

\begin{table*}[h]\footnotesize
	\centering
	\caption{The cost of transferring of propositions of evidences}
		\begin{tabular}{c c c c| c c  c c c c c c c c c c}\hline
			$Evidences$ & \multicolumn{3}{c}{$Costs \ \ for \ \ proposition\ \ a $}&\multicolumn{3}{c}{$Costs \ \ for \ \ proposition\ \ b $} \\\hline
			$Evidence_{1}$&$\{a \rightarrow a\}$& $\{a \rightarrow b\}$ & $\{a \rightarrow c\}$& $\{b \rightarrow a\}$ &$\{b \rightarrow b\}$& $\{b \rightarrow c\}$  \\
			& $0.5000 $ & $0.9241 $ & $0.7311 $
			& $0.0759 $ & $0.5000 $ & $0.1824 $\\\hline
			& \multicolumn{3}{c}{$Costs \ \ for \ \ proposition\ \ c $} \\\hline
			& $\{c \rightarrow a\}$ & $\{c \rightarrow b\}$ &$\{c \rightarrow c\}$ \\
			& $0.2689 $ & $0.8176 $ & $0.5000 $\\\hline
			& \multicolumn{3}{c}{$Costs \ \ for \ \ proposition\ \ a $}&\multicolumn{3}{c}{$Costs \ \ for \ \ proposition\ \ b $} \\\hline
			$Evidence_{2}$&$\{a \rightarrow a\}$& $\{a \rightarrow b\}$ & $\{a \rightarrow c\}$& $\{b \rightarrow a\}$ &$\{b \rightarrow b\}$& $\{b \rightarrow c\}$  \\
			& $0.5000 $ & $0.3775 $ & $0.9241 $ & $0.6225 $ & $0.5000 $ & $0.9526 $
			\\\hline
			
			& \multicolumn{3}{c}{$Costs \ \ for \ \ proposition\ \ c $} \\\hline
			& $\{c \rightarrow a\}$ & $\{c \rightarrow b\}$ &$\{c \rightarrow c\}$ \\
			& $0.0759 $ & $0.0474 $ & $0.5000 $\\\hline	
			& \multicolumn{3}{c}{$Costs \ \ for \ \ proposition\ \ a $}&\multicolumn{3}{c}{$Costs \ \ for \ \ proposition\ \ b $} \\\hline
			$Evidence_{3}$&$\{a \rightarrow a\}$& $\{a \rightarrow b\}$ & $\{a \rightarrow c\}$& $\{b \rightarrow a\}$ &$\{b \rightarrow b\}$& $\{b \rightarrow c\}$  \\
			& $0.5000 $ & $0.2689 $ & $0.1824 $ & $0.7311 $ & $0.5000 $ & $0.3775 $\\\hline
			& \multicolumn{3}{c}{$Costs \ \ for \ \ proposition\ \ c $} \\\hline
			& $\{c \rightarrow a\}$ & $\{c \rightarrow b\}$ &$\{c \rightarrow c\}$ \\
			& $0.8176 $ & $0.6225 $ & $0.5000 $\\\hline
			
		\end{tabular}
	
	\label{tableuu}
\end{table*}

\begin{table*}[h]\footnotesize
	\centering
	\caption{The TPs of transferring of propositions of evidences}
		\begin{tabular}{c c c c| c c  c c c c c c c c c c}\hline
			$Evidences$ & \multicolumn{3}{c}{$TPs \ \ for \ \ proposition\ \ a $}&\multicolumn{3}{c}{$TPs \ \ for \ \ proposition\ \ b $} \\\hline
			$Evidence_{1}$&$\{a \rightarrow a\}$& $\{a \rightarrow b\}$ & $\{a \rightarrow c\}$& $\{b \rightarrow a\}$ &$\{b \rightarrow b\}$& $\{b \rightarrow c\}$  \\
			& $0.2320 $ & $0.4288 $ & $0.3392 $ & $0.1000 $ & $0.6594 $ & $0.2406 $\\\hline
			& \multicolumn{3}{c}{$TPs \ \ for \ \ proposition\ \ c $} \\\hline
			& $\{c \rightarrow a\}$ & $\{c \rightarrow b\}$ &$\{c \rightarrow c\}$ \\
			& $0.1695 $ & $0.5153 $ & $0.3152 $\\\hline
			& \multicolumn{3}{c}{$TPs \ \ for \ \ proposition\ \ a $}&\multicolumn{3}{c}{$TPs \ \ for \ \ proposition\ \ b $} \\\hline
			$Evidence_{2}$&$\{a \rightarrow a\}$& $\{a \rightarrow b\}$ & $\{a \rightarrow c\}$& $\{b \rightarrow a\}$ &$\{b \rightarrow b\}$& $\{b \rightarrow c\}$  \\
			& $0.2775 $ & $0.2095 $ & $0.5129 $ & $0.3000 $ & $0.2410 $ & $0.4591 $\\\hline
			
			& \multicolumn{3}{c}{$TPs \ \ for \ \ proposition\ \ c $} \\\hline
			& $\{c \rightarrow a\}$ & $\{c \rightarrow b\}$ &$\{c \rightarrow c\}$ \\
			& $0.1217 $ & $0.0761 $ & $0.8022 $\\\hline	
			& \multicolumn{3}{c}{$TPs \ \ for \ \ proposition\ \ a $}&\multicolumn{3}{c}{$TPs \ \ for \ \ proposition\ \ b $} \\\hline
			$Evidence_{3}$&$\{a \rightarrow a\}$& $\{a \rightarrow b\}$ & $\{a \rightarrow c\}$& $\{b \rightarrow a\}$ &$\{b \rightarrow b\}$& $\{b \rightarrow c\}$  \\
			& $0.5256 $ & $0.2827 $ & $0.1918 $ & $0.4545 $ & $0.3108 $ & $0.2347 $\\\hline
			& \multicolumn{3}{c}{$TPs \ \ for \ \ proposition\ \ c $} \\\hline
			& $\{c \rightarrow a\}$ & $\{c \rightarrow b\}$ &$\{c \rightarrow c\}$ \\
			& $0.4214 $ & $0.3208 $ & $0.2577 $\\\hline
		\end{tabular}
	
	\label{table33}
\end{table*}

\begin{table*}[h]\footnotesize
	\centering
	\caption{The results of $d_{MC}$}
		\begin{tabular}{c c c c c c  c c c c c c c c c c}\hline
			$Distance\ d_{MC}$ & $Evidence_{1}^{MC} \leftrightarrow Evidence_{1}^{MC}$ &$Evidence_{1}^{MC} \leftrightarrow Evidence_{2}^{MC}$\\
			&$0$ &$0.4582$ \\\hline
			&$Evidence_{1}^{MC} \leftrightarrow Evidence_{3}^{MC}$\\
			&$0.3605$\\\hline
			$Distance\ d_{MC}$ & $Evidence_{2}^{MC} \leftrightarrow Evidence_{1}^{MC}$ &$Evidence_{2}^{MC} \leftrightarrow Evidence_{2}^{MC}$\\
			&$0.4582$ &$0$ \\\hline
			&$Evidence_{2}^{MC} \leftrightarrow Evidence_{3}^{MC}$\\
			&$0.4358$\\\hline
			$Distance\ d_{MC}$ & $Evidence_{3}^{MC} \leftrightarrow Evidence_{1}^{MC}$ &$Evidence_{3}^{MC} \leftrightarrow Evidence_{2}^{MC}$\\
			&$0.3605$ &$0.4358$ \\\hline
			&$Evidence_{3}^{MC} \leftrightarrow Evidence_{3}^{MC}$\\
			&$0$\\\hline
			
		\end{tabular}
	
	\label{table334}
\end{table*}

\begin{table*}[h]\footnotesize
	\centering
	\caption{The results of $d_{TP}$}
		\begin{tabular}{c c c c c c  c c c c c c c c c c}\hline
			$Distance\ d_{TP}$ & $Evidence_{1}^{TP} \leftrightarrow Evidence_{1}^{TP}$ &$Evidence_{1}^{TP} \leftrightarrow Evidence_{2}^{TP}$\\
			&$0$ &$1.0278$ \\\hline
			&$Evidence_{1}^{TP} \leftrightarrow Evidence_{3}^{TP}$\\
			&$0.8345$\\\hline
			$Distance\ d_{TP}$ & $Evidence_{2}^{TP} \leftrightarrow Evidence_{1}^{TP}$ &$Evidence_{2}^{TP} \leftrightarrow Evidence_{2}^{TP}$\\
			&$1.0278$ &$0$ \\\hline
			&$Evidence_{2}^{TP} \leftrightarrow Evidence_{3}^{TP}$\\
			&$0.9627$\\\hline
			$Distance\ d_{TP}$ & $Evidence_{3}^{TP} \leftrightarrow Evidence_{1}^{TP}$ &$Evidence_{3}^{TP} \leftrightarrow Evidence_{2}^{TP}$\\
			&$0.8345$ &$0.9627$ \\\hline
			&$Evidence_{3}^{TP} \leftrightarrow Evidence_{3}^{TP}$\\
			&$0$\\\hline
		\end{tabular}
	
	\label{table335}
\end{table*}

\begin{table*}[h]\footnotesize
	\centering
	\caption{The results of modified evidences}
		\begin{tabular}{c c c c c c  c c c c c c c c c c}\hline
			$Evidences$ & $Evidence_{1}^{Modified}$ &$Evidence_{2}^{Modified}$&$Evidence_{3}^{Modified}$\\
			& $0.0298 $ & $0.8307 $ & $0.1395 $ \\\hline
			$Evidences$ & $Evidence_{1}^{Modified}$ &$Evidence_{2}^{Modified}$&$Evidence_{3}^{Modified}$\\
			& $0.0425 $ & $0.0172 $ & $0.9403 $ \\\hline
			$Evidences$ & $Evidence_{1}^{Modified}$ &$Evidence_{2}^{Modified}$&$Evidence_{3}^{Modified}$\\
			& $0.6690 $ & $0.2156 $ & $0.1154 $ \\\hline
		\end{tabular}
	
	\label{table338}
\end{table*}

\begin{table*}[h]\footnotesize
	\centering
	\caption{The results of $d_{ME}$}
		\begin{tabular}{c c c c| c c  c c c c c c c c c c}\hline
			$Distance\ d_{ME}$ & $Evidence_{1}^{ME} \leftrightarrow Evidence_{1}^{ME}$ &$Evidence_{1}^{ME} \leftrightarrow Evidence_{2}^{ME}$\\
			&$0$ &$0.8072$ \\\hline
			&$Evidence_{1}^{ME} \leftrightarrow Evidence_{3}^{ME}$\\
			&$0.6274$\\\hline
			$Distance\ d_{ME}$ & $Evidence_{2}^{ME} \leftrightarrow Evidence_{1}^{ME}$ &$Evidence_{2}^{ME} \leftrightarrow Evidence_{2}^{ME}$\\
			&$0.8072$ &$0$ \\\hline
			&$Evidence_{2}^{ME} \leftrightarrow Evidence_{3}^{ME}$\\
			&$0.7457$\\\hline
			$Distance\ d_{ME}$ & $Evidence_{1}^{ME} \leftrightarrow Evidence_{3}^{ME}$ &$Evidence_{3}^{ME} \leftrightarrow Evidence_{2}^{ME}$\\
			&$0.6274$ &$0.7457$ \\\hline
			&$Evidence_{3}^{ME} \leftrightarrow Evidence_{3}^{ME}$\\
			&$0$\\\hline
		\end{tabular}
	
	\label{table336}
\end{table*}

\begin{table}[h]\footnotesize
	\centering
	\caption{The results of similarity measure}
		\begin{tabular}{c c c c c c  c c c c c c c c c c}\hline
			Proposed method&\multicolumn{3}{c}{$Values \ \ of \ \ SM $}\\\hline
			& $Evidence_{1}$ &$Evidence_{2}$&$Evidence_{3}$\\
			&$3.0419$ &$2.8213$ &$3.1562$\\\hline
			Deng $et \ al.$&\multicolumn{3}{c}{$Values \ \ of \ \ similarity $}\\\hline
			& $Evidence_{1}$ &$Evidence_{2}$&$Evidence_{3}$\\
			&$0.3383$ &$ 0.3168$ &$0.3448$\\\hline
		\end{tabular}
	
	\label{table337}
\end{table}

In fact, the original intention to design the algorithm to obtain the degree of similarity is to take all of the features and underlying influential factors contained in evidences into consideration, which may eliminate some errors caused by partial deviations of the values of propositions. For example, if one kind of measure of distance produces counter-intuitive results, but the rest of the measure of distance on other series of data can probably offset the negative effect produced by one category of measure. In this example, it can be obtained that the similarity measure is able to generate consistent judgment with the ones produced by Deng $et \ al.$'s method. However, considering the degree of divergence of the three evidences provided, for decision maker, a targeted decision can not be made and it is better think every evidence owns a similar confidential support. At this point, the proposed method performs better by assigning a similar confidential degree to each of evidence considering the much bigger base number and difference among the value of judgment. All in all, the proposed method possesses all the advantages of Deng $et \ al.$'s method and can resist the effects brought by some individual abnormal results. For the proposed method, a much more steady and reliable results of judgments is the most contribution of the proposed method.

\textbf{Example: }Elimination for abnormal values, improved combination and certainty measure

\subsection{The basic condition of the evidence group}
Assume 16 pieces of evidences in GET are given by sensors. The values of propositions of evidences are listed in Table \ref{table1}. 
\begin{table*}[h]\footnotesize
	\centering
	\caption{The evidences given by sensors}
			\begin{tabular}{c c c c c c c c c c c c c c c c}\hline
				$Evidences$ & \multicolumn{5}{c}{$Values \ \ of \ \ propositions $}&$Evidences$ & \multicolumn{5}{c}{$Values \ \ of \ \ propositions $}\\\hline
				$Evidence_{1}$& $\{a\}$ & $\{b\}$ &$\{c\}$&$\{d\}$&$\{\emptyset\}$&$Evidence_{2}$& $\{a\}$ & $\{b\}$ &$\{c\}$&$\{d\}$&$\{\emptyset\}$ \\
				& $0.3$ & $0.34$&$0.09$&$0.13$&$0.14$ && $0.32$ & $0.38$&$0.14$&$0.07$&$0.09$\\
				$Evidence_{3}$& $\{a\}$ & $\{b\}$ &$\{c\}$&$\{d\}$&$\{\emptyset\}$&$Evidence_{4}$& $\{a\}$ & $\{b\}$ &$\{c\}$&$\{d\}$&$\{\emptyset\}$ \\
				& $0.29$ & $0.36$&$0.13$&$0.1$&$0.12$ && $0.22$ & $0.25$&$0.31$&$0.02$&$0.2$\\
				$Evidence_{5}$& $\{a\}$ & $\{b\}$ &$\{c\}$&$\{d\}$&$\{\emptyset\}$&$Evidence_{6}$& $\{a\}$ & $\{b\}$ &$\{c\}$&$\{d\}$&$\{\emptyset\}$ \\
				& $0.31$ & $0.39$&$0.11$&$0.09$&$0.1$ && $0.33$ & $0.35$&$0.12$&$0.09$&$0.11$\\
				$Evidence_{7}$& $\{a\}$ & $\{b\}$ &$\{c\}$&$\{d\}$&$\{\emptyset\}$&$Evidence_{8}$& $\{a\}$ & $\{b\}$ &$\{c\}$&$\{d\}$&$\{\emptyset\}$ \\
				& $0.33$ & $0.21$&$0.16$&$0.18$&$0.12$ && $0.312$ & $0.368$&$0.177$&$0.13$&$0.013$\\
				$Evidence_{9}$& $\{a\}$ & $\{b\}$ &$\{c\}$&$\{d\}$&$\{\emptyset\}$&$Evidence_{10}$& $\{a\}$ & $\{b\}$ &$\{c\}$&$\{d\}$&$\{\emptyset\}$ \\
				& $0.31$ & $0.345$&$0.122$&$0.132$&$0.091$ && $0.343$ & $0.354$&$0.12$&$0.111$&$0.072$\\
				$Evidence_{11}$& $\{a\}$ & $\{b\}$ &$\{c\}$&$\{d\}$&$\{\emptyset\}$&$Evidence_{12}$& $\{a\}$ & $\{b\}$ &$\{c\}$&$\{d\}$&$\{\emptyset\}$ \\
				& $0.318$ & $0.37$&$0.13$&$0.1$&$0.082$ && $0.34$ & $0.322$&$0.127$&$0.125$&$0.086$\\
				$Evidence_{13}$& $\{a\}$ & $\{b\}$ &$\{c\}$&$\{d\}$&$\{\emptyset\}$&$Evidence_{14}$& $\{a\}$ & $\{b\}$ &$\{c\}$&$\{d\}$&$\{\emptyset\}$ \\
				& $0.326$ & $0.346$&$0.124$&$0.11$&$0.094$ && $0.347$ & $0.339$&$0.16$&$0.095$&$0.059$\\
				$Evidence_{15}$& $\{a\}$ & $\{b\}$ &$\{c\}$&$\{d\}$&$\{\emptyset\}$&$Evidence_{16}$& $\{a\}$ & $\{b\}$ &$\{c\}$&$\{d\}$&$\{\emptyset\}$ \\
				& $0.047$ & $0.218$&$0.11$&$0.185$&$0.44$ && $0.324$ & $0.357$&$0.071$&$0.132$&$0.116$\\
				\hline
		\end{tabular}
	
	\label{table1}
\end{table*}

\subsection{The management using normal distribution}

\begin{figure}[h]
	\centering
	\includegraphics[scale=0.4]{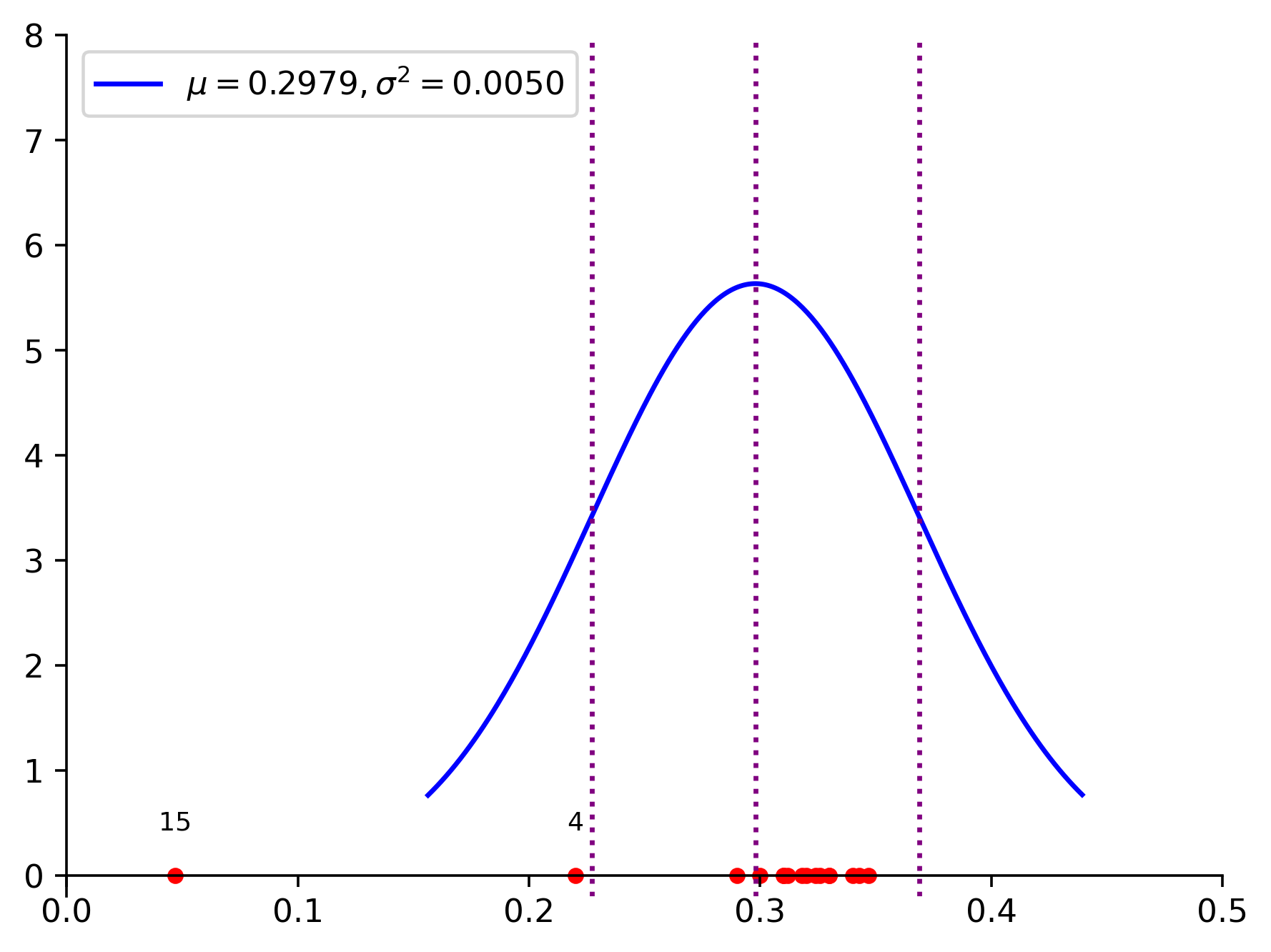}
	\caption{The situations of proposition $a$}
	\label{opop}
\end{figure}

\begin{figure}[h]
	\centering
	\includegraphics[scale=0.4]{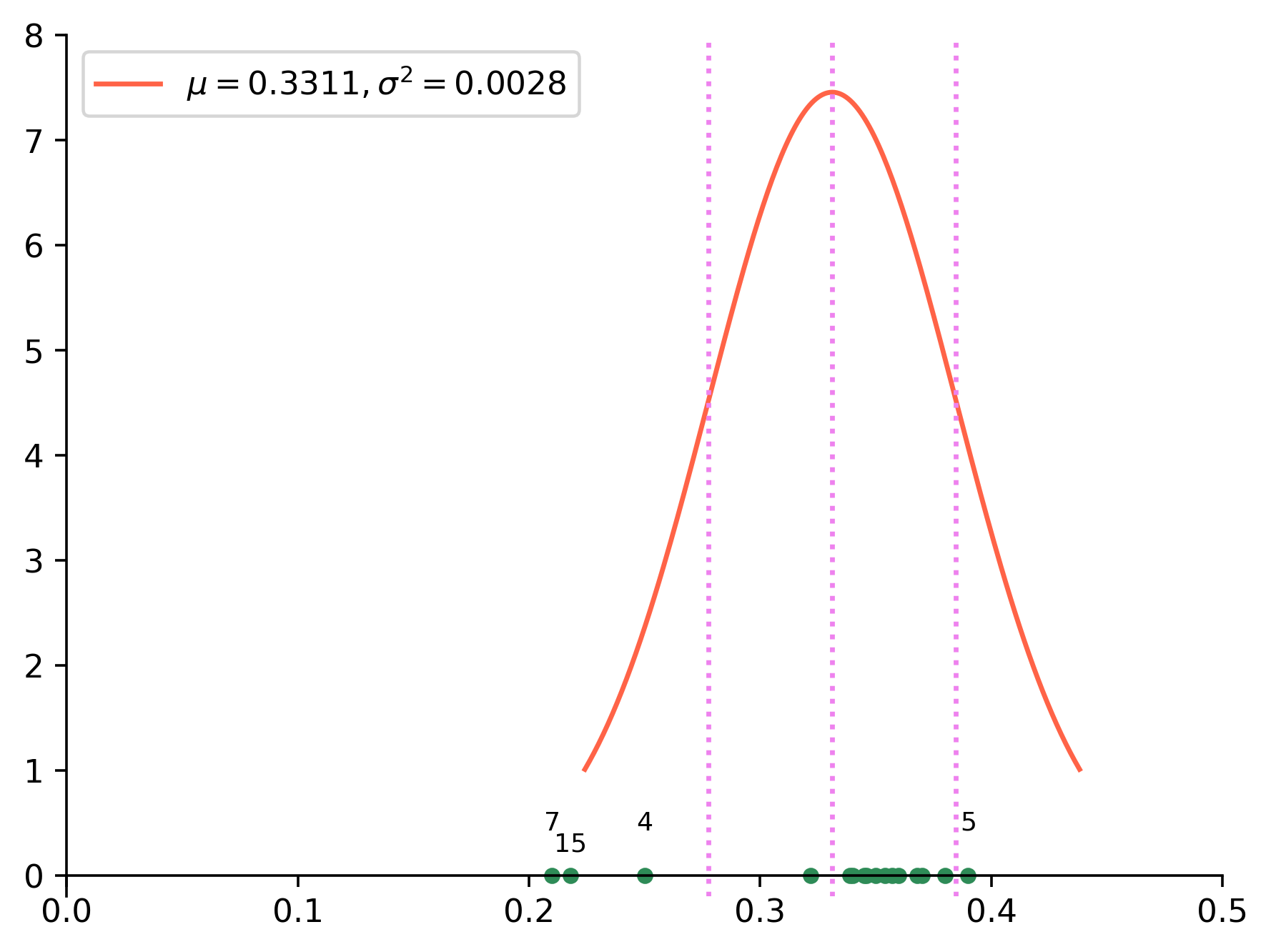}
	\caption{The situations of proposition $b$}
	\label{opop1}
\end{figure}

\begin{figure}[h]
	\centering
	\includegraphics[scale=0.4]{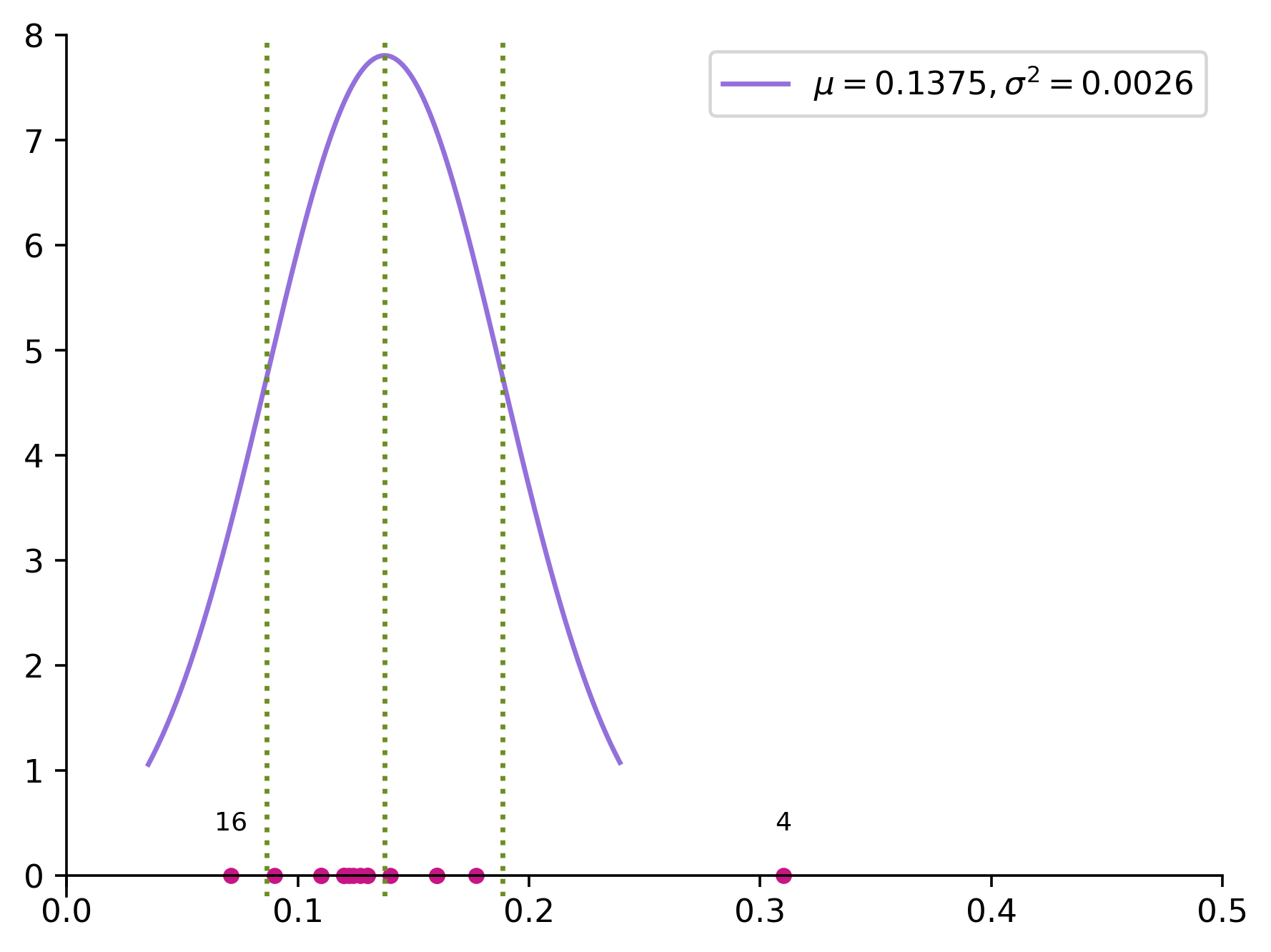}
	\caption{The situations of proposition $c$}
	\label{opopp}
\end{figure}

\begin{figure}[h]
	\centering
	\includegraphics[scale=0.4]{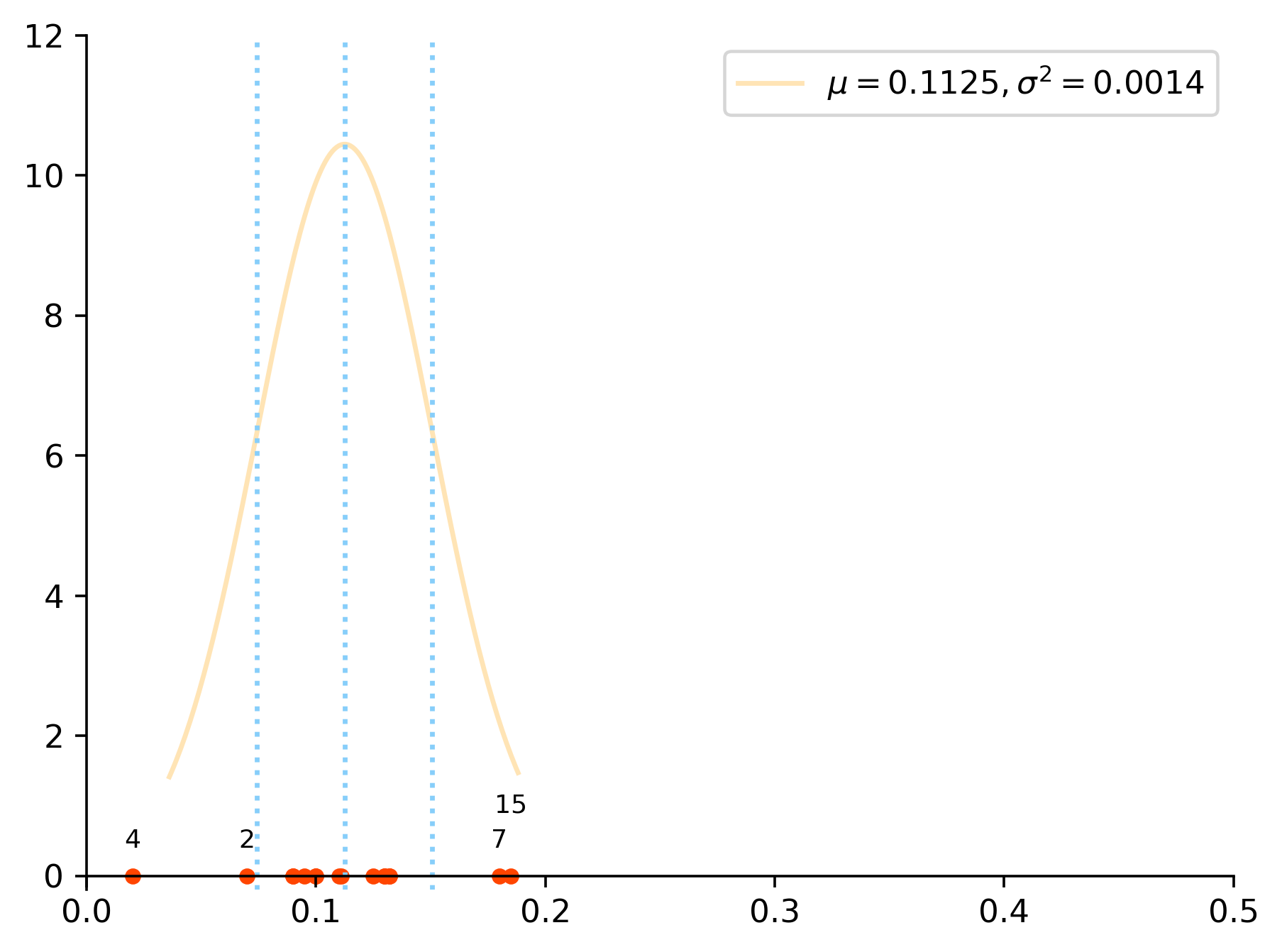}
	\caption{The situations of proposition $d$}
	\label{opopps}
\end{figure}

\begin{figure}[h]
	\centering
	\includegraphics[scale=0.4]{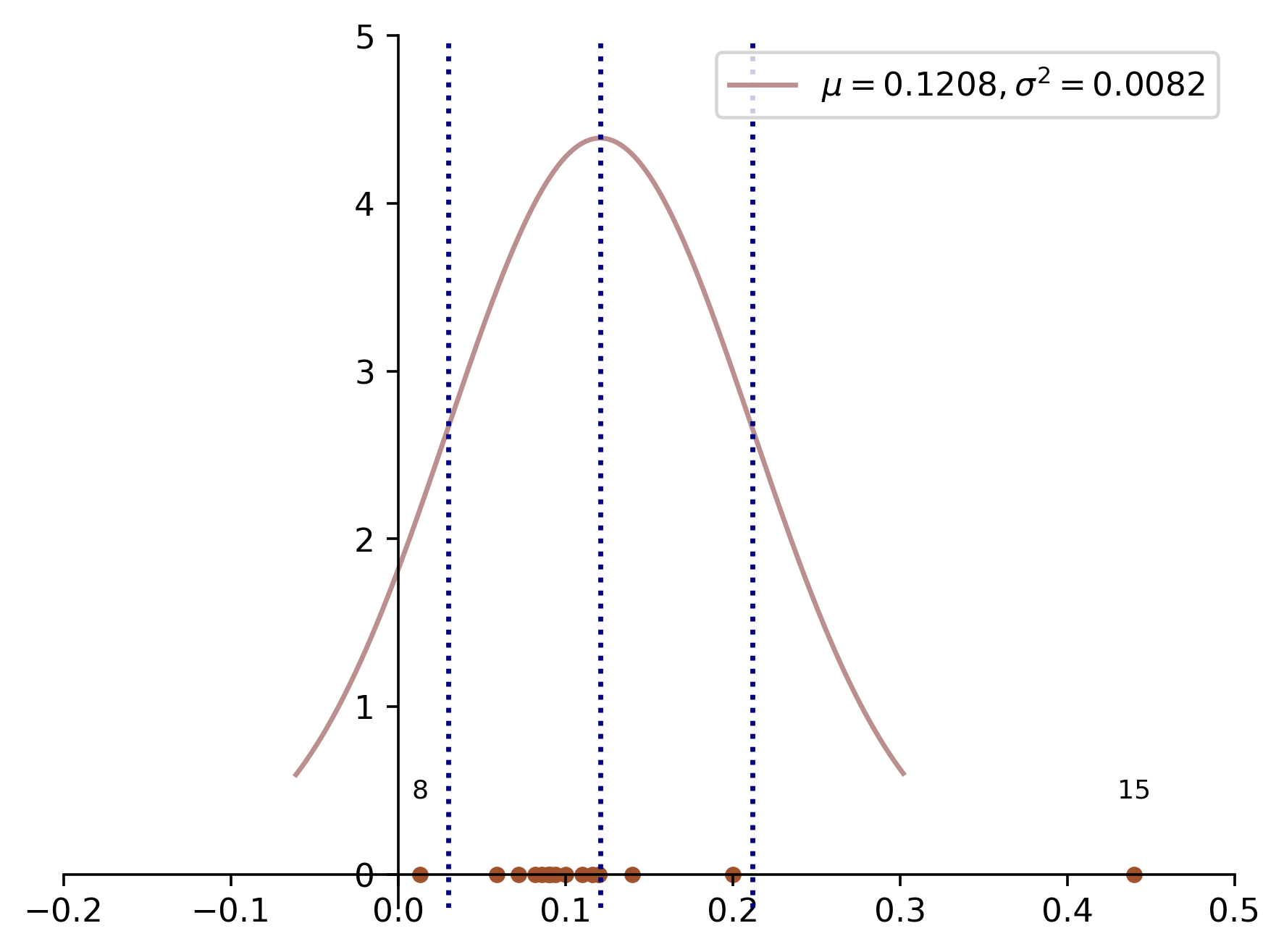}
	\caption{The situations of proposition $\emptyset$}
	\label{Fig11}
\end{figure}

Then, a model based on normal distribution is constructed and the process is shown in Figure \ref{opop}, \ref{opop1}, \ref{opopp}, \ref{opopps} and \ref{Fig11}.

Through the step of elimination of the informal values, the 2nd, 4th, 5th, 7th, 8th, 15th and 16th piece of evidences is erased. By checking the values of propositions,it can be concluded that the abnormal values of some propositions are accurately recognized. This operation successfully alleviates conflicts contained among evidences to avoid producing counter-intuitive results of judgement on the practical situations.

\subsection{The results of calculation of the transition cost and probability}
Based on the results obtained from the operation of normal distribution, the cost for each proposition in each evidence are calculated being given in Table \ref{table3}, \ref{table4}. Besides, according to the definition of CM, the variable CM can be obtained by utilizing the TPs corresponding to the specific propositions. TPs are also listed in Table \ref{table5} and \ref{table6}. By analysing the data of the cost, it can be easily concluded that if the state of the proposition intends to transfers to itself, the cost is  assigned a unitive value $0.5000$, which gives a continuous description on the transition of the original state without changes among different states. And in general, the values indicates that, when a state with a lower mass attempts to transfer to a state with a higher mass, the cost rises dramatically, which is corresponding to the theory of increase of entropy. On the contrary, the cost decreases to indicate everything owns a trend to become uncertain. Therefore, the TPs obtained based on its definition can actually reflect the situations of all of the process of transition. The phenomenon can be simply summarised that the if the state transfers from a proposition which owns a bigger possibility to take place to a smaller one, the according probability is homologous lower and the corresponding probability is relatively higher in the opposite circumstances. All in all, the results produced conforms to the actual situations and agree with the intuitive judgments.

\begin{table*}[h]\footnotesize
	\centering
	\caption{The first group of cost of transferring of propositions of evidences}
			\begin{tabular}{c c c c c c | c c c c c c c c c c}\hline
				$Evidences$ & \multicolumn{5}{c}{$Costs \ \ for \ \ proposition\ \ a $}&\multicolumn{5}{c}{$Costs \ \ for \ \ proposition\ \ b $} \\\hline
				$Evidence_{1}$&$\{a \rightarrow a\}$& $\{a \rightarrow b\}$ & $\{a \rightarrow c\}$ &$\{a \rightarrow d\}$ & $\{a \rightarrow \emptyset\}$& $\{b \rightarrow a\}$ &$\{b \rightarrow b\}$& $\{b \rightarrow c\}$ &$\{b \rightarrow d\}$ & $\{b \rightarrow \emptyset\}$ \\
				& $0.5000$ & $0.5498$& $0.2592$ & $0.2994$&$0.3100$&$0.4501$&$0.5000$ & $0.2227$ & $0.2592$&$0.2689$\\\hline
				& \multicolumn{5}{c}{$Costs \ \ for \ \ proposition\ \ c $}&\multicolumn{5}{c}{$Costs \ \ for \ \ proposition\ \ d $} \\\hline
				& $\{c \rightarrow a\}$ & $\{c \rightarrow b\}$ &$\{c \rightarrow c\}$&$\{c \rightarrow d\}$ & $\{c \rightarrow \emptyset\}$& $\{d \rightarrow a\}$ & $\{d \rightarrow b\}$ &$\{d \rightarrow c\}$&$\{d \rightarrow d\}$ & $\{d \rightarrow \emptyset\}$ \\
				& $0.7407$ & $0.7772$& $0.5000$ & $0.5498$&$0.5621$&$0.7005$&$0.7407$ & $0.4501$ & $0.5000$&$0.5124$\\\hline
				& \multicolumn{5}{c}{$Costs \ \ for \ \ proposition\ \ \emptyset $} \\\hline
				& $\{\emptyset \rightarrow a\}$ &$\{\emptyset \rightarrow b\}$& $\{\emptyset \rightarrow c\}$ &$\{\emptyset \rightarrow d\}$ & $\{\emptyset \rightarrow \emptyset\}$\\
				& $0.6899$& $0.7310$ & $0.4378$&$0.4875$&$0.5000$\\\hline
				$Evidence_{3}$&$\{a \rightarrow a\}$& $\{a \rightarrow b\}$ & $\{a \rightarrow c\}$ &$\{a \rightarrow d\}$ & $\{a \rightarrow \emptyset\}$& $\{b \rightarrow a\}$ &$\{b \rightarrow b\}$& $\{b \rightarrow c\}$ &$\{b \rightarrow d\}$ & $\{b \rightarrow \emptyset\}$ \\
				& $0.5000$ & $0.5866$& $0.3100$ & $0.2788$&$0.2994$&$0.4133$&$0.5000$ & $0.2404$ & $0.2141$&$0.2314$\\\hline
				& \multicolumn{5}{c}{$Costs \ \ for \ \ proposition\ \ c $}&\multicolumn{5}{c}{$Costs \ \ for \ \ proposition\ \ d $} \\\hline
				& $\{c \rightarrow a\}$ & $\{c \rightarrow b\}$ &$\{c \rightarrow c\}$&$\{c \rightarrow d\}$ & $\{c \rightarrow \emptyset\}$& $\{d \rightarrow a\}$ & $\{d \rightarrow b\}$ &$\{d \rightarrow c\}$&$\{d \rightarrow d\}$ & $\{d \rightarrow \emptyset\}$ \\
				& $0.6899$ & $0.7595$& $0.5000$ & $0.4625$&$0.4875$&$0.7211$&$0.7858$ & $0.5374$ & $0.5000$&$0.5249$\\\hline
				& \multicolumn{5}{c}{$Costs \ \ for \ \ proposition\ \ \emptyset $} \\\hline
				& $\{\emptyset \rightarrow a\}$ &$\{\emptyset \rightarrow b\}$& $\{\emptyset \rightarrow c\}$ &$\{\emptyset \rightarrow d\}$ & $\{\emptyset \rightarrow \emptyset\}$\\
				& $0.7005$& $0.7685$ & $0.5124$&$0.4750$&$0.5000$\\\hline
				$Evidence_{6}$&$\{a \rightarrow a\}$& $\{a \rightarrow b\}$ & $\{a \rightarrow c\}$ &$\{a \rightarrow d\}$ & $\{a \rightarrow \emptyset\}$& $\{b \rightarrow a\}$ &$\{b \rightarrow b\}$& $\{b \rightarrow c\}$ &$\{b \rightarrow d\}$ & $\{b \rightarrow \emptyset\}$ \\
				& $0.5000 $ & $0.5250 $ & $0.2592 $ & $0.2315 $ & $0.2497 $& $0.4750 $ & $0.5000 $ & $0.2405 $ & $0.2142 $ & $0.2315 $\\\hline
				& \multicolumn{5}{c}{$Costs \ \ for \ \ proposition\ \ c $}&\multicolumn{5}{c}{$Costs \ \ for \ \ proposition\ \ d $} \\\hline
				& $\{c \rightarrow a\}$ & $\{c \rightarrow b\}$ &$\{c \rightarrow c\}$&$\{c \rightarrow d\}$ & $\{c \rightarrow \emptyset\}$& $\{d \rightarrow a\}$ & $\{d \rightarrow b\}$ &$\{d \rightarrow c\}$&$\{d \rightarrow d\}$ & $\{d \rightarrow \emptyset\}$ \\
				& $0.7408 $ & $0.7595 $ & $0.5000 $ & $0.4626 $ & $0.4875 $& $0.7685 $ & $0.7858 $ & $0.5374 $ & $0.5000 $ & $0.5250 $\\\hline
				& \multicolumn{5}{c}{$Costs \ \ for \ \ proposition\ \ \emptyset $} \\\hline
				& $\{\emptyset \rightarrow a\}$ &$\{\emptyset \rightarrow b\}$& $\{\emptyset \rightarrow c\}$ &$\{\emptyset \rightarrow d\}$ & $\{\emptyset \rightarrow \emptyset\}$\\
				& $0.7503 $ & $0.7685 $ & $0.5125 $ & $0.4750 $ & $0.5000 $\\\hline
				$Evidence_{9}$&$\{a \rightarrow a\}$& $\{a \rightarrow b\}$ & $\{a \rightarrow c\}$ &$\{a \rightarrow d\}$ & $\{a \rightarrow \emptyset\}$& $\{b \rightarrow a\}$ &$\{b \rightarrow b\}$& $\{b \rightarrow c\}$ &$\{b \rightarrow d\}$ & $\{b \rightarrow \emptyset\}$ \\
				& $0.5000 $ & $0.5436 $ & $0.2809 $ & $0.2911 $ & $0.2507 $& $0.4564 $ & $0.5000 $ & $0.2469 $ & $0.2564 $ & $0.2193 $\\\hline
				& \multicolumn{5}{c}{$Costs \ \ for \ \ proposition\ \ c $}&\multicolumn{5}{c}{$Costs \ \ for \ \ proposition\ \ d $} \\\hline
				& $\{c \rightarrow a\}$ & $\{c \rightarrow b\}$ &$\{c \rightarrow c\}$&$\{c \rightarrow d\}$ & $\{c \rightarrow \emptyset\}$& $\{d \rightarrow a\}$ & $\{d \rightarrow b\}$ &$\{d \rightarrow c\}$&$\{d \rightarrow d\}$ & $\{d \rightarrow \emptyset\}$ \\
				& $0.7191 $ & $0.7531 $ & $0.5000 $ & $0.5125 $ & $0.4613 $& $0.7089 $ & $0.7436 $ & $0.4875 $ & $0.5000 $ & $0.4489 $\\\hline
				& \multicolumn{5}{c}{$Costs \ \ for \ \ proposition\ \ \emptyset $} \\\hline
				& $\{\emptyset \rightarrow a\}$ &$\{\emptyset \rightarrow b\}$& $\{\emptyset \rightarrow c\}$ &$\{\emptyset \rightarrow d\}$ & $\{\emptyset \rightarrow \emptyset\}$\\
				& $0.7493 $ & $0.7807 $ & $0.5387 $ & $0.5511 $ & $0.5000 $\\\hline
		\end{tabular}
	
	\label{table3}
\end{table*}

\begin{table*}[h]\footnotesize
	\centering
	\caption{The second group of cost of transferring of propositions of evidences}
			\begin{tabular}{c c c c c c | c c c c c c c c c c}\hline
				$Evidences$ & \multicolumn{5}{c}{$Costs \ \ for \ \ proposition\ \ a $}&\multicolumn{5}{c}{$Costs \ \ for \ \ proposition\ \ b $} \\\hline
				$Evidence_{10}$&$\{a \rightarrow a\}$& $\{a \rightarrow b\}$ & $\{a \rightarrow c\}$ &$\{a \rightarrow d\}$ & $\{a \rightarrow \emptyset\}$& $\{b \rightarrow a\}$ &$\{b \rightarrow b\}$& $\{b \rightarrow c\}$ &$\{b \rightarrow d\}$ & $\{b \rightarrow \emptyset\}$ \\
				& $0.5000 $ & $0.5137 $ & $0.2469 $ & $0.2387 $ & $0.2051 $
				& $0.4863 $ & $0.5000 $ & $0.2369 $ & $0.2288 $ & $0.1962 $\\\hline
				& \multicolumn{5}{c}{$Costs \ \ for \ \ proposition\ \ c $}&\multicolumn{5}{c}{$Costs \ \ for \ \ proposition\ \ d $} \\\hline
				& $\{c \rightarrow a\}$ & $\{c \rightarrow b\}$ &$\{c \rightarrow c\}$&$\{c \rightarrow d\}$ & $\{c \rightarrow \emptyset\}$& $\{d \rightarrow a\}$ & $\{d \rightarrow b\}$ &$\{d \rightarrow c\}$&$\{d \rightarrow d\}$ & $\{d \rightarrow \emptyset\}$ \\
				& $0.7531 $ & $0.7631 $ & $0.5000 $ & $0.4888 $ & $0.4403 $
				
				& $0.7613 $ & $0.7712 $ & $0.5112 $ & $0.5000 $ & $0.4514 $
				\\\hline
				& \multicolumn{5}{c}{$Costs \ \ for \ \ proposition\ \ \emptyset $} \\\hline
				& $\{\emptyset \rightarrow a\}$ &$\{\emptyset \rightarrow b\}$& $\{\emptyset \rightarrow c\}$ &$\{\emptyset \rightarrow d\}$ & $\{\emptyset \rightarrow \emptyset\}$\\
				& $0.7949 $ & $0.8038 $ & $0.5597 $ & $0.5486 $ & $0.5000 $\\\hline
				$Evidence_{11}$&$\{a \rightarrow a\}$& $\{a \rightarrow b\}$ & $\{a \rightarrow c\}$ &$\{a \rightarrow d\}$ & $\{a \rightarrow \emptyset\}$& $\{b \rightarrow a\}$ &$\{b \rightarrow b\}$& $\{b \rightarrow c\}$ &$\{b \rightarrow d\}$ & $\{b \rightarrow \emptyset\}$ \\
				& $0.5000 $ & $0.5646 $ & $0.2809 $ & $0.2516 $ & $0.2351 $
				
				& $0.4354 $ & $0.5000 $ & $0.2315 $ & $0.2059 $ & $0.1915 $\\\hline
				& \multicolumn{5}{c}{$Costs \ \ for \ \ proposition\ \ c $}&\multicolumn{5}{c}{$Costs \ \ for \ \ proposition\ \ d $} \\\hline
				& $\{c \rightarrow a\}$ & $\{c \rightarrow b\}$ &$\{c \rightarrow c\}$&$\{c \rightarrow d\}$ & $\{c \rightarrow \emptyset\}$& $\{d \rightarrow a\}$ & $\{d \rightarrow b\}$ &$\{d \rightarrow c\}$&$\{d \rightarrow d\}$ & $\{d \rightarrow \emptyset\}$ \\
				& $0.7191 $ & $0.7685 $ & $0.5000 $ & $0.4626 $ & $0.4403 $
				
				& $0.7484 $ & $0.7941 $ & $0.5374 $ & $0.5000 $ & $0.4775 $\\\hline
				& \multicolumn{5}{c}{$Costs \ \ for \ \ proposition\ \ \emptyset $} \\\hline
				& $\{\emptyset \rightarrow a\}$ &$\{\emptyset \rightarrow b\}$& $\{\emptyset \rightarrow c\}$ &$\{\emptyset \rightarrow d\}$ & $\{\emptyset \rightarrow \emptyset\}$\\
				& $0.7649 $ & $0.8085 $ & $0.5597 $ & $0.5225 $ & $0.5000 $\\\hline
				$Evidence_{12}$&$\{a \rightarrow a\}$& $\{a \rightarrow b\}$ & $\{a \rightarrow c\}$ &$\{a \rightarrow d\}$ & $\{a \rightarrow \emptyset\}$& $\{b \rightarrow a\}$ &$\{b \rightarrow b\}$& $\{b \rightarrow c\}$ &$\{b \rightarrow d\}$ & $\{b \rightarrow \emptyset\}$ \\
				& $0.5000 $ & $0.4775 $ & $0.2564 $ & $0.2545 $ & $0.2193 $
				
				& $0.5225 $ & $0.5000 $ & $0.2739 $ & $0.2719 $ & $0.2351 $\\\hline
				& \multicolumn{5}{c}{$Costs \ \ for \ \ proposition\ \ c $}&\multicolumn{5}{c}{$Costs \ \ for \ \ proposition\ \ d $} \\\hline
				& $\{c \rightarrow a\}$ & $\{c \rightarrow b\}$ &$\{c \rightarrow c\}$&$\{c \rightarrow d\}$ & $\{c \rightarrow \emptyset\}$& $\{d \rightarrow a\}$ & $\{d \rightarrow b\}$ &$\{d \rightarrow c\}$&$\{d \rightarrow d\}$ & $\{d \rightarrow \emptyset\}$ \\
				& $0.7436 $ & $0.7261 $ & $0.5000 $ & $0.4975 $ & $0.4489 $
				
				& $0.7455 $ & $0.7281 $ & $0.5025 $ & $0.5000 $ & $0.4514 $\\\hline
				& \multicolumn{5}{c}{$Costs \ \ for \ \ proposition\ \ \emptyset $} \\\hline
				& $\{\emptyset \rightarrow a\}$ &$\{\emptyset \rightarrow b\}$& $\{\emptyset \rightarrow c\}$ &$\{\emptyset \rightarrow d\}$ & $\{\emptyset \rightarrow \emptyset\}$\\
				& $0.7807 $ & $0.7649 $ & $0.5511 $ & $0.5486 $ & $0.5000 $\\\hline
				$Evidence_{13}$&$\{a \rightarrow a\}$& $\{a \rightarrow b\}$ & $\{a \rightarrow c\}$ &$\{a \rightarrow d\}$ & $\{a \rightarrow \emptyset\}$& $\{b \rightarrow a\}$ &$\{b \rightarrow b\}$& $\{b \rightarrow c\}$ &$\{b \rightarrow d\}$ & $\{b \rightarrow \emptyset\}$ \\
				& $0.5000 $ & $0.5250 $ & $0.2670 $ & $0.2535 $ & $0.2387 $
				
				& $0.4750 $ & $0.5000 $ & $0.2479 $ & $0.2351 $ & $0.2210 $
				\\\hline
				& \multicolumn{5}{c}{$Costs \ \ for \ \ proposition\ \ c $}&\multicolumn{5}{c}{$Costs \ \ for \ \ proposition\ \ d $} \\\hline
				& $\{c \rightarrow a\}$ & $\{c \rightarrow b\}$ &$\{c \rightarrow c\}$&$\{c \rightarrow d\}$ & $\{c \rightarrow \emptyset\}$& $\{d \rightarrow a\}$ & $\{d \rightarrow b\}$ &$\{d \rightarrow c\}$&$\{d \rightarrow d\}$ & $\{d \rightarrow \emptyset\}$ \\
				& $0.7330 $ & $0.7521 $ & $0.5000 $ & $0.4825 $ & $0.4626 $
				
				& $0.7465 $ & $0.7649 $ & $0.5175 $ & $0.5000 $ & $0.4800 $\\\hline
				& \multicolumn{5}{c}{$Costs \ \ for \ \ proposition\ \ \emptyset $} \\\hline
				& $\{\emptyset \rightarrow a\}$ &$\{\emptyset \rightarrow b\}$& $\{\emptyset \rightarrow c\}$ &$\{\emptyset \rightarrow d\}$ & $\{\emptyset \rightarrow \emptyset\}$\\
				& $0.7613 $ & $0.7790 $ & $0.5374 $ & $0.5200 $ & $0.5000 $\\\hline
				$Evidence_{14}$&$\{a \rightarrow a\}$& $\{a \rightarrow b\}$ & $\{a \rightarrow c\}$ &$\{a \rightarrow d\}$ & $\{a \rightarrow \emptyset\}$& $\{b \rightarrow a\}$ &$\{b \rightarrow b\}$& $\{b \rightarrow c\}$ &$\{b \rightarrow d\}$ & $\{b \rightarrow \emptyset\}$ \\
				& $0.5000 $ & $0.4900 $ & $0.2819 $ & $0.2210 $ & $0.1915 $
				
				& $0.5100 $ & $0.5000 $ & $0.2901 $ & $0.2279 $ & $0.1978 $\\\hline
				& \multicolumn{5}{c}{$Costs \ \ for \ \ proposition\ \ c $}&\multicolumn{5}{c}{$Costs \ \ for \ \ proposition\ \ d $} \\\hline
				& $\{c \rightarrow a\}$ & $\{c \rightarrow b\}$ &$\{c \rightarrow c\}$&$\{c \rightarrow d\}$ & $\{c \rightarrow \emptyset\}$& $\{d \rightarrow a\}$ & $\{d \rightarrow b\}$ &$\{d \rightarrow c\}$&$\{d \rightarrow d\}$ & $\{d \rightarrow \emptyset\}$ \\
				& $0.7181 $ & $0.7099 $ & $0.5000 $ & $0.4195 $ & $0.3764 $
				
				& $0.7790 $ & $0.7721 $ & $0.5805 $ & $0.5000 $ & $0.4551 $
				\\\hline
				& \multicolumn{5}{c}{$Costs \ \ for \ \ proposition\ \ \emptyset $} \\\hline
				& $\{\emptyset \rightarrow a\}$ &$\{\emptyset \rightarrow b\}$& $\{\emptyset \rightarrow c\}$ &$\{\emptyset \rightarrow d\}$ & $\{\emptyset \rightarrow \emptyset\}$\\
				& $0.8085 $ & $0.8022 $ & $0.6236 $ & $0.5449 $ & $0.5000 $\\\hline
		\end{tabular}
	
	\label{table4}
\end{table*}

\begin{table*}[h]\footnotesize
	\centering
	\caption{The first group of TPs of transferring of propositions of evidences}
			\begin{tabular}{c c c c c c | c c c c c c c c c c}\hline
				$Evidences$ & \multicolumn{5}{c}{$TPs \ \ for \ \ proposition\ \ a $}&\multicolumn{5}{c}{$TPs \ \ for \ \ proposition\ \ b $} \\\hline
				$Evidence_{1}$&$\{a \rightarrow a\}$& $\{a \rightarrow b\}$ & $\{a \rightarrow c\}$ &$\{a \rightarrow d\}$ & $\{a \rightarrow \emptyset\}$& $\{b \rightarrow a\}$ &$\{b \rightarrow b\}$& $\{b \rightarrow c\}$ &$\{b \rightarrow d\}$ & $\{b \rightarrow \emptyset\}$ \\
				& $0.2606 $ & $0.2866 $ & $0.1351 $ & $0.1561 $ & $0.1616 $
				
				& $0.2646 $ & $0.2939 $ & $0.1309 $ & $0.1524 $ & $0.1581 $\\\hline
				& \multicolumn{5}{c}{$TPs \ \ for \ \ proposition\ \ c $}&\multicolumn{5}{c}{$TPs \ \ for \ \ proposition\ \ d $} \\\hline
				& $\{c \rightarrow a\}$ & $\{c \rightarrow b\}$ &$\{c \rightarrow c\}$&$\{c \rightarrow d\}$ & $\{c \rightarrow \emptyset\}$& $\{d \rightarrow a\}$ & $\{d \rightarrow b\}$ &$\{d \rightarrow c\}$&$\{d \rightarrow d\}$ & $\{d \rightarrow \emptyset\}$ \\
				& $0.2367 $ & $0.2483 $ & $0.1597 $ & $0.1757 $ & $0.1796 $
				
				& $0.2412 $ & $0.2551 $ & $0.1550 $ & $0.1722 $ & $0.1765 $
				\\\hline
				& \multicolumn{5}{c}{$TPs \ \ for \ \ proposition\ \ \emptyset $} \\\hline
				& $\{\emptyset \rightarrow a\}$ &$\{\emptyset \rightarrow b\}$& $\{\emptyset \rightarrow c\}$ &$\{\emptyset \rightarrow d\}$ & $\{\emptyset \rightarrow \emptyset\}$\\
				& $0.2424 $ & $0.2568 $ & $0.1538 $ & $0.1713 $ & $0.1757 $\\\hline
				$Evidence_{3}$&$\{a \rightarrow a\}$& $\{a \rightarrow b\}$ & $\{a \rightarrow c\}$ &$\{a \rightarrow d\}$ & $\{a \rightarrow \emptyset\}$& $\{b \rightarrow a\}$ &$\{b \rightarrow b\}$& $\{b \rightarrow c\}$ &$\{b \rightarrow d\}$ & $\{b \rightarrow \emptyset\}$ \\
				& $0.2532 $ & $0.2970 $ & $0.1570 $ & $0.1412 $ & $0.1516 $
				
				& $0.2584 $ & $0.3126 $ & $0.1504 $ & $0.1339 $ & $0.1447 $\\\hline
				& \multicolumn{5}{c}{$TPs \ \ for \ \ proposition\ \ c $}&\multicolumn{5}{c}{$TPs \ \ for \ \ proposition\ \ d $} \\\hline
				& $\{c \rightarrow a\}$ & $\{c \rightarrow b\}$ &$\{c \rightarrow c\}$&$\{c \rightarrow d\}$ & $\{c \rightarrow \emptyset\}$& $\{d \rightarrow a\}$ & $\{d \rightarrow b\}$ &$\{d \rightarrow c\}$&$\{d \rightarrow d\}$ & $\{d \rightarrow \emptyset\}$ \\
				& $0.2380 $ & $0.2619 $ & $0.1724 $ & $0.1595 $ & $0.1681 $
				
				& $0.2349 $ & $0.2560 $ & $0.1751 $ & $0.1629 $ & $0.1710 $
				\\\hline
				& \multicolumn{5}{c}{$TPs \ \ for \ \ proposition\ \ \emptyset $} \\\hline
				& $\{\emptyset \rightarrow a\}$ &$\{\emptyset \rightarrow b\}$& $\{\emptyset \rightarrow c\}$ &$\{\emptyset \rightarrow d\}$ & $\{\emptyset \rightarrow \emptyset\}$\\
				& $0.2369 $ & $0.2599 $ & $0.1733 $ & $0.1607 $ & $0.1691 $\\\hline
				$Evidence_{6}$&$\{a \rightarrow a\}$& $\{a \rightarrow b\}$ & $\{a \rightarrow c\}$ &$\{a \rightarrow d\}$ & $\{a \rightarrow \emptyset\}$& $\{b \rightarrow a\}$ &$\{b \rightarrow b\}$& $\{b \rightarrow c\}$ &$\{b \rightarrow d\}$ & $\{b \rightarrow \emptyset\}$ \\
				& $0.2832 $ & $0.2974 $ & $0.1468 $ & $0.1311 $ & $0.1415 $
				
				& $0.2860 $ & $0.3010 $ & $0.1448 $ & $0.1289 $ & $0.1393 $\\\hline
				& \multicolumn{5}{c}{$TPs \ \ for \ \ proposition\ \ c $}&\multicolumn{5}{c}{$TPs \ \ for \ \ proposition\ \ d $} \\\hline
				& $\{c \rightarrow a\}$ & $\{c \rightarrow b\}$ &$\{c \rightarrow c\}$&$\{c \rightarrow d\}$ & $\{c \rightarrow \emptyset\}$& $\{d \rightarrow a\}$ & $\{d \rightarrow b\}$ &$\{d \rightarrow c\}$&$\{d \rightarrow d\}$ & $\{d \rightarrow \emptyset\}$ \\
				& $0.2511 $ & $0.2574 $ & $0.1695 $ & $0.1568 $ & $0.1652 $
				
				& $0.2466 $ & $0.2521 $ & $0.1724 $ & $0.1604 $ & $0.1684 $
				\\\hline
				& \multicolumn{5}{c}{$TPs \ \ for \ \ proposition\ \ \emptyset $} \\\hline
				& $\{\emptyset \rightarrow a\}$ &$\{\emptyset \rightarrow b\}$& $\{\emptyset \rightarrow c\}$ &$\{\emptyset \rightarrow d\}$ & $\{\emptyset \rightarrow \emptyset\}$\\
				& $0.2496 $ & $0.2556 $ & $0.1705 $ & $0.1580 $ & $0.1663 $\\\hline
				$Evidence_{9}$&$\{a \rightarrow a\}$& $\{a \rightarrow b\}$ & $\{a \rightarrow c\}$ &$\{a \rightarrow d\}$ & $\{a \rightarrow \emptyset\}$& $\{b \rightarrow a\}$ &$\{b \rightarrow b\}$& $\{b \rightarrow c\}$ &$\{b \rightarrow d\}$ & $\{b \rightarrow \emptyset\}$ \\
				& $0.2679 $ & $0.2913 $ & $0.1505 $ & $0.1560 $ & $0.1343 $
				
				& $0.2718 $ & $0.2978 $ & $0.1471 $ & $0.1527 $ & $0.1306 $\\\hline
				& \multicolumn{5}{c}{$TPs \ \ for \ \ proposition\ \ c $}&\multicolumn{5}{c}{$TPs \ \ for \ \ proposition\ \ d $} \\\hline
				& $\{c \rightarrow a\}$ & $\{c \rightarrow b\}$ &$\{c \rightarrow c\}$&$\{c \rightarrow d\}$ & $\{c \rightarrow \emptyset\}$& $\{d \rightarrow a\}$ & $\{d \rightarrow b\}$ &$\{d \rightarrow c\}$&$\{d \rightarrow d\}$ & $\{d \rightarrow \emptyset\}$ \\
				& $0.2441 $ & $0.2556 $ & $0.1697 $ & $0.1740 $ & $0.1566 $
				
				& $0.2454 $ & $0.2574 $ & $0.1687 $ & $0.1731 $ & $0.1554 $
				\\\hline
				& \multicolumn{5}{c}{$TPs \ \ for \ \ proposition\ \ \emptyset $} \\\hline
				& $\{\emptyset \rightarrow a\}$ &$\{\emptyset \rightarrow b\}$& $\{\emptyset \rightarrow c\}$ &$\{\emptyset \rightarrow d\}$ & $\{\emptyset \rightarrow \emptyset\}$\\
				& $0.2402 $ & $0.2503 $ & $0.1727 $ & $0.1766 $ & $0.1603 $\\\hline
		\end{tabular}
	
	\label{table5}
\end{table*}

\begin{table*}[h]\footnotesize
	\centering
	\caption{The second group of TPs of transferring of propositions of evidences}

			\begin{tabular}{c c c c c c | c c c c c c c c c c}\hline
				$Evidences$ & \multicolumn{5}{c}{$TPs \ \ for \ \ proposition\ \ a $}&\multicolumn{5}{c}{$TPs \ \ for \ \ proposition\ \ b $} \\\hline
				$Evidence_{10}$&$\{a \rightarrow a\}$& $\{a \rightarrow b\}$ & $\{a \rightarrow c\}$ &$\{a \rightarrow d\}$ & $\{a \rightarrow \emptyset\}$& $\{b \rightarrow a\}$ &$\{b \rightarrow b\}$& $\{b \rightarrow c\}$ &$\{b \rightarrow d\}$ & $\{b \rightarrow \emptyset\}$ \\
				& $0.2934 $ & $0.3014 $ & $0.1449 $ & $0.1400 $ & $0.1203 $
				
				& $0.2950 $ & $0.3034 $ & $0.1437 $ & $0.1388 $ & $0.1191 $\\\hline
				& \multicolumn{5}{c}{$TPs \ \ for \ \ proposition\ \ c $}&\multicolumn{5}{c}{$TPs \ \ for \ \ proposition\ \ d $} \\\hline
				& $\{c \rightarrow a\}$ & $\{c \rightarrow b\}$ &$\{c \rightarrow c\}$&$\{c \rightarrow d\}$ & $\{c \rightarrow \emptyset\}$& $\{d \rightarrow a\}$ & $\{d \rightarrow b\}$ &$\{d \rightarrow c\}$&$\{d \rightarrow d\}$ & $\{d \rightarrow \emptyset\}$ \\
				& $0.2557 $ & $0.2591 $ & $0.1698 $ & $0.1659 $ & $0.1495 $
				
				& $0.2542 $ & $0.2575 $ & $0.1707 $ & $0.1669 $ & $0.1507 $
				\\\hline
				& \multicolumn{5}{c}{$TPs \ \ for \ \ proposition\ \ \emptyset $} \\\hline
				& $\{\emptyset \rightarrow a\}$ &$\{\emptyset \rightarrow b\}$& $\{\emptyset \rightarrow c\}$ &$\{\emptyset \rightarrow d\}$ & $\{\emptyset \rightarrow \emptyset\}$\\
				& $0.2479 $ & $0.2506 $ & $0.1745 $ & $0.1711 $ & $0.1559 $\\\hline
				$Evidence_{11}$&$\{a \rightarrow a\}$& $\{a \rightarrow b\}$ & $\{a \rightarrow c\}$ &$\{a \rightarrow d\}$ & $\{a \rightarrow \emptyset\}$& $\{b \rightarrow a\}$ &$\{b \rightarrow b\}$& $\{b \rightarrow c\}$ &$\{b \rightarrow d\}$ & $\{b \rightarrow \emptyset\}$ \\
				& $0.2729 $ & $0.3082 $ & $0.1533 $ & $0.1373 $ & $0.1283 $
				
				& $0.2783 $ & $0.3196 $ & $0.1480 $ & $0.1316 $ & $0.1225 $\\\hline
				& \multicolumn{5}{c}{$TPs \ \ for \ \ proposition\ \ c $}&\multicolumn{5}{c}{$TPs \ \ for \ \ proposition\ \ d $} \\\hline
				& $\{c \rightarrow a\}$ & $\{c \rightarrow b\}$ &$\{c \rightarrow c\}$&$\{c \rightarrow d\}$ & $\{c \rightarrow \emptyset\}$& $\{d \rightarrow a\}$ & $\{d \rightarrow b\}$ &$\{d \rightarrow c\}$&$\{d \rightarrow d\}$ & $\{d \rightarrow \emptyset\}$ \\
				& $0.2488 $ & $0.2659 $ & $0.1730 $ & $0.1600 $ & $0.1523 $
				
				& $0.2448 $ & $0.2597 $ & $0.1758 $ & $0.1635 $ & $0.1562 $\\\hline
				& \multicolumn{5}{c}{$TPs \ \ for \ \ proposition\ \ \emptyset $} \\\hline
				& $\{\emptyset \rightarrow a\}$ &$\{\emptyset \rightarrow b\}$& $\{\emptyset \rightarrow c\}$ &$\{\emptyset \rightarrow d\}$ & $\{\emptyset \rightarrow \emptyset\}$\\
				& $0.2424 $ & $0.2562 $ & $0.1774 $ & $0.1656 $ & $0.1584 $\\\hline
				$Evidence_{12}$&$\{a \rightarrow a\}$& $\{a \rightarrow b\}$ & $\{a \rightarrow c\}$ &$\{a \rightarrow d\}$ & $\{a \rightarrow \emptyset\}$& $\{b \rightarrow a\}$ &$\{b \rightarrow b\}$& $\{b \rightarrow c\}$ &$\{b \rightarrow d\}$ & $\{b \rightarrow \emptyset\}$ \\
				& $0.2928 $ & $0.2796 $ & $0.1501 $ & $0.1490 $ & $0.1284 $
				
				& $0.2897 $ & $0.2773 $ & $0.1519 $ & $0.1508 $ & $0.1303 $
				\\\hline
				& \multicolumn{5}{c}{$TPs \ \ for \ \ proposition\ \ c $}&\multicolumn{5}{c}{$TPs \ \ for \ \ proposition\ \ d $} \\\hline
				& $\{c \rightarrow a\}$ & $\{c \rightarrow b\}$ &$\{c \rightarrow c\}$&$\{c \rightarrow d\}$ & $\{c \rightarrow \emptyset\}$& $\{d \rightarrow a\}$ & $\{d \rightarrow b\}$ &$\{d \rightarrow c\}$&$\{d \rightarrow d\}$ & $\{d \rightarrow \emptyset\}$ \\
				& $0.2550 $ & $0.2490 $ & $0.1715 $ & $0.1706 $ & $0.1539 $
				
				& $0.2547 $ & $0.2487 $ & $0.1716 $ & $0.1708 $ & $0.1542 $\\\hline
				& \multicolumn{5}{c}{$TPs \ \ for \ \ proposition\ \ \emptyset $} \\\hline
				& $\{\emptyset \rightarrow a\}$ &$\{\emptyset \rightarrow b\}$& $\{\emptyset \rightarrow c\}$ &$\{\emptyset \rightarrow d\}$ & $\{\emptyset \rightarrow \emptyset\}$\\
				& $0.2482 $ & $0.2432 $ & $0.1752 $ & $0.1744 $ & $0.1590 $\\\hline
				$Evidence_{13}$&$\{a \rightarrow a\}$& $\{a \rightarrow b\}$ & $\{a \rightarrow c\}$ &$\{a \rightarrow d\}$ & $\{a \rightarrow \emptyset\}$& $\{b \rightarrow a\}$ &$\{b \rightarrow b\}$& $\{b \rightarrow c\}$ &$\{b \rightarrow d\}$ & $\{b \rightarrow \emptyset\}$ \\
				& $0.2802 $ & $0.2942 $ & $0.1496 $ & $0.1421 $ & $0.1338 $
				
				& $0.2829 $ & $0.2978 $ & $0.1476 $ & $0.1400 $ & $0.1316 $
				\\\hline
				& \multicolumn{5}{c}{$TPs \ \ for \ \ proposition\ \ c $}&\multicolumn{5}{c}{$TPs \ \ for \ \ proposition\ \ d $} \\\hline
				& $\{c \rightarrow a\}$ & $\{c \rightarrow b\}$ &$\{c \rightarrow c\}$&$\{c \rightarrow d\}$ & $\{c \rightarrow \emptyset\}$& $\{d \rightarrow a\}$ & $\{d \rightarrow b\}$ &$\{d \rightarrow c\}$&$\{d \rightarrow d\}$ & $\{d \rightarrow \emptyset\}$ \\
				& $0.2502 $ & $0.2567 $ & $0.1706 $ & $0.1647 $ & $0.1579 $
				
				& $0.2481 $ & $0.2542 $ & $0.1720 $ & $0.1662 $ & $0.1595 $\\\hline
				& \multicolumn{5}{c}{$TPs \ \ for \ \ proposition\ \ \emptyset $} \\\hline
				& $\{\emptyset \rightarrow a\}$ &$\{\emptyset \rightarrow b\}$& $\{\emptyset \rightarrow c\}$ &$\{\emptyset \rightarrow d\}$ & $\{\emptyset \rightarrow \emptyset\}$\\
				& $0.2458 $ & $0.2515 $ & $0.1735 $ & $0.1679 $ & $0.1614 $\\\hline
				$Evidence_{14}$&$\{a \rightarrow a\}$& $\{a \rightarrow b\}$ & $\{a \rightarrow c\}$ &$\{a \rightarrow d\}$ & $\{a \rightarrow \emptyset\}$& $\{b \rightarrow a\}$ &$\{b \rightarrow b\}$& $\{b \rightarrow c\}$ &$\{b \rightarrow d\}$ & $\{b \rightarrow \emptyset\}$ \\
				& $0.2968 $ & $0.2909 $ & $0.1674 $ & $0.1312 $ & $0.1137 $
				
				& $0.2955 $ & $0.2897 $ & $0.1681 $ & $0.1321 $ & $0.1146 $\\\hline
				& \multicolumn{5}{c}{$TPs \ \ for \ \ proposition\ \ c $}&\multicolumn{5}{c}{$TPs \ \ for \ \ proposition\ \ d $} \\\hline
				& $\{c \rightarrow a\}$ & $\{c \rightarrow b\}$ &$\{c \rightarrow c\}$&$\{c \rightarrow d\}$ & $\{c \rightarrow \emptyset\}$& $\{d \rightarrow a\}$ & $\{d \rightarrow b\}$ &$\{d \rightarrow c\}$&$\{d \rightarrow d\}$ & $\{d \rightarrow \emptyset\}$ \\
				& $0.2636 $ & $0.2606 $ & $0.1836 $ & $0.1540 $ & $0.1382 $
				
				& $0.2524 $ & $0.2501 $ & $0.1881 $ & $0.1620 $ & $0.1474 $
				\\\hline
				& \multicolumn{5}{c}{$TPs \ \ for \ \ proposition\ \ \emptyset $} \\\hline
				& $\{\emptyset \rightarrow a\}$ &$\{\emptyset \rightarrow b\}$& $\{\emptyset \rightarrow c\}$ &$\{\emptyset \rightarrow d\}$ & $\{\emptyset \rightarrow \emptyset\}$\\
				& $0.2465 $ & $0.2446 $ & $0.1902 $ & $0.1662 $ & $0.1525 $\\\hline
		\end{tabular}
	
	\label{table6}
\end{table*}

\subsection{The results of CMs}

\begin{figure}[h]
	\centering
	\includegraphics[scale=0.4]{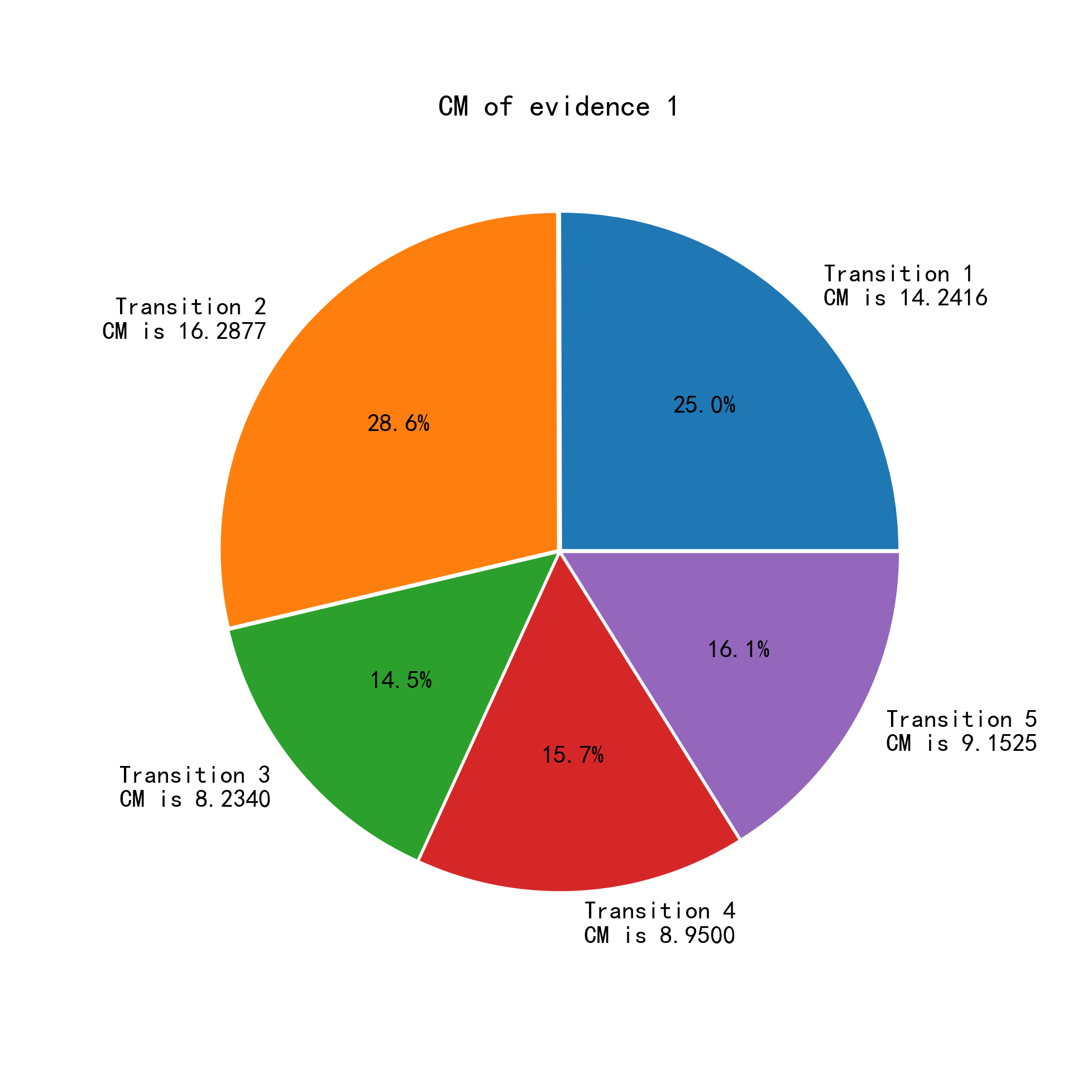}
	\caption{The proportion of different transitions in evidence 1}
	\label{1}
\end{figure}

\begin{figure}[h]
	\centering
	\includegraphics[scale=0.4]{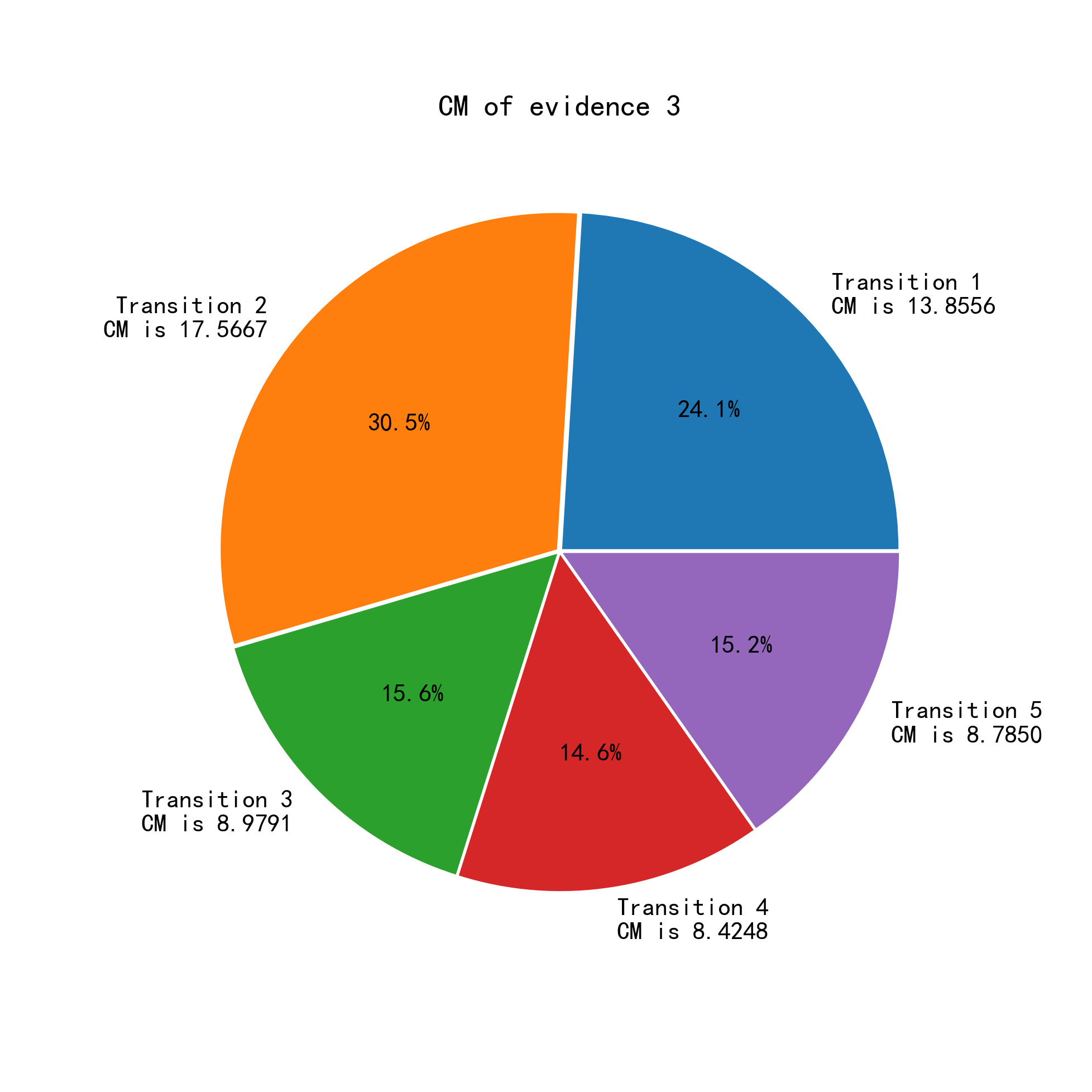}
	\caption{The proportion of different transitions in evidence 3}
	\label{2}
\end{figure}

\begin{figure}[h]
	\centering
	\includegraphics[scale=0.4]{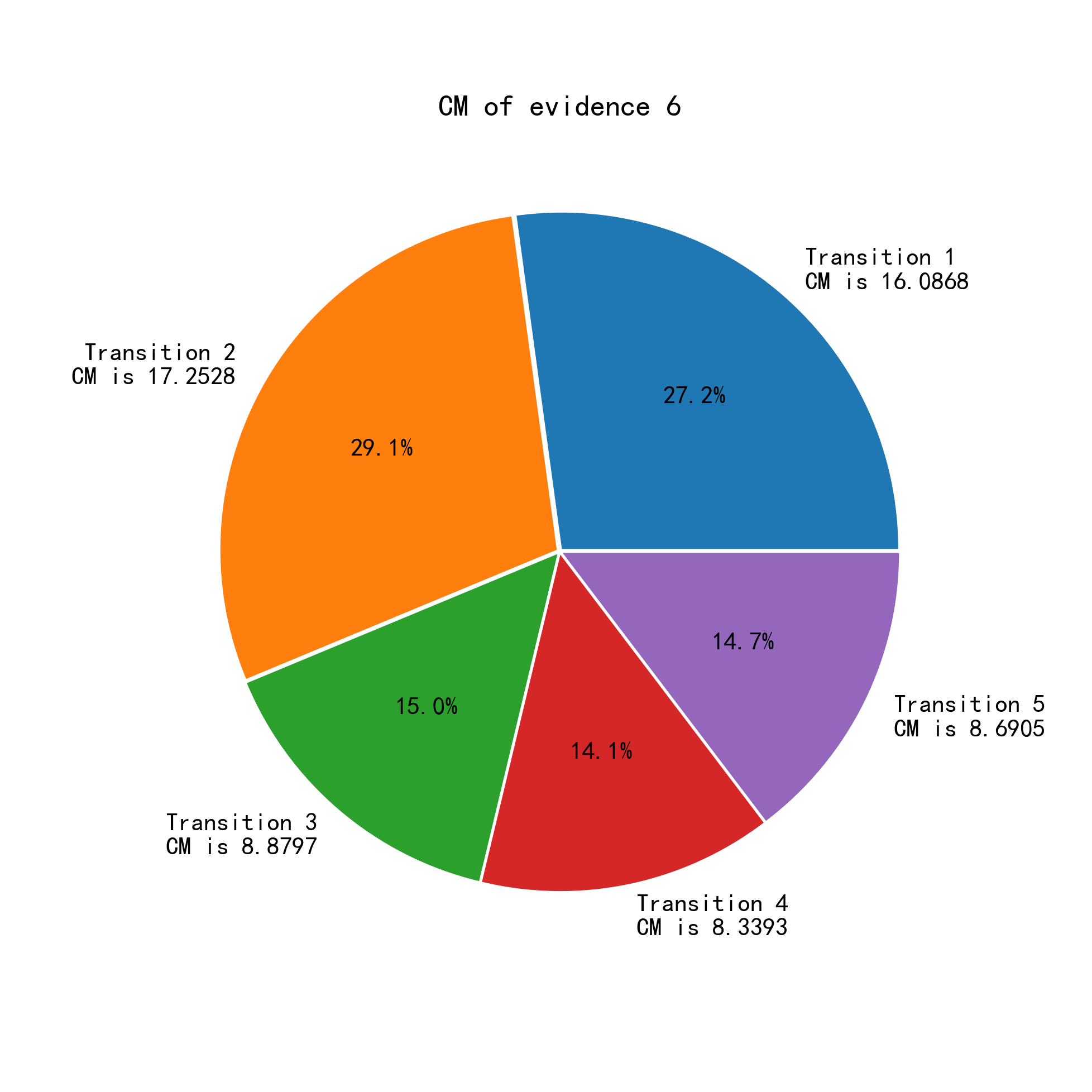}
	\caption{The proportion of different transitions in evidence 6}
	\label{3}
\end{figure}

\begin{figure}[h]
	\centering
	\includegraphics[scale=0.4]{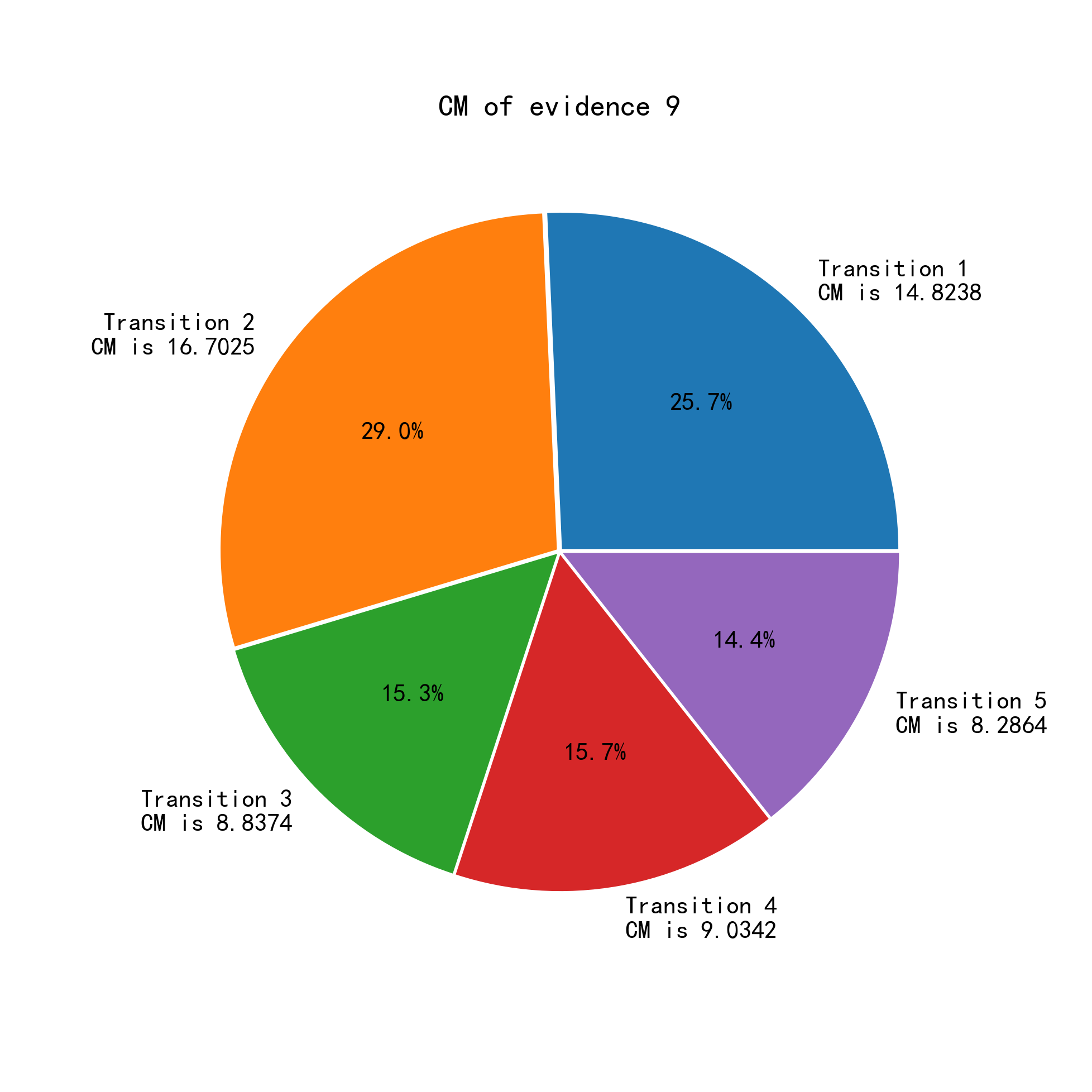}
	\caption{The proportion of different transitions in evidence 9}
	\label{4}
\end{figure}

\begin{figure}[h]
	\centering
	\includegraphics[scale=0.4]{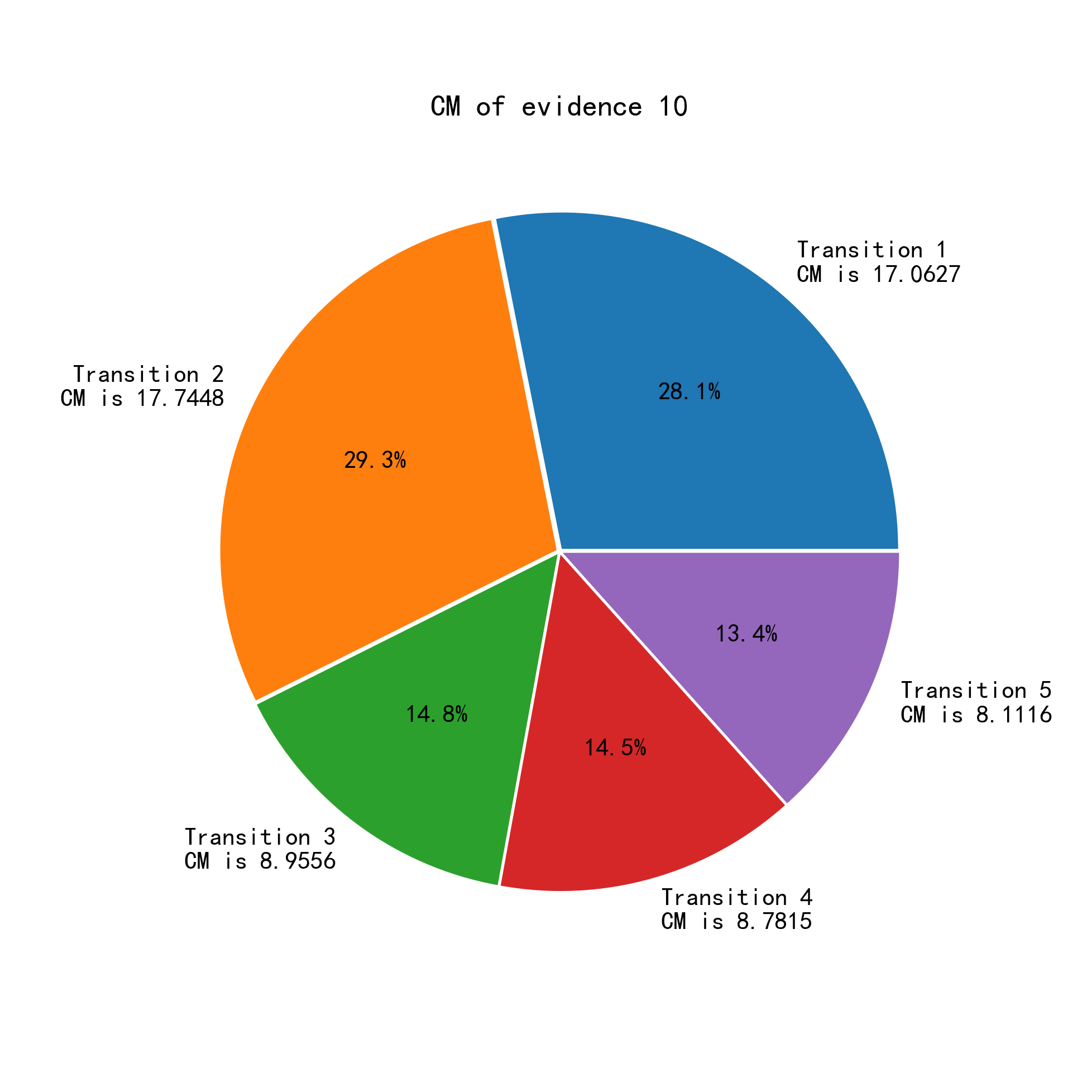}
	\caption{The proportion of different transitions in evidence 10}
	\label{5}
\end{figure}

\begin{figure}[h]
	\centering
	\includegraphics[scale=0.4]{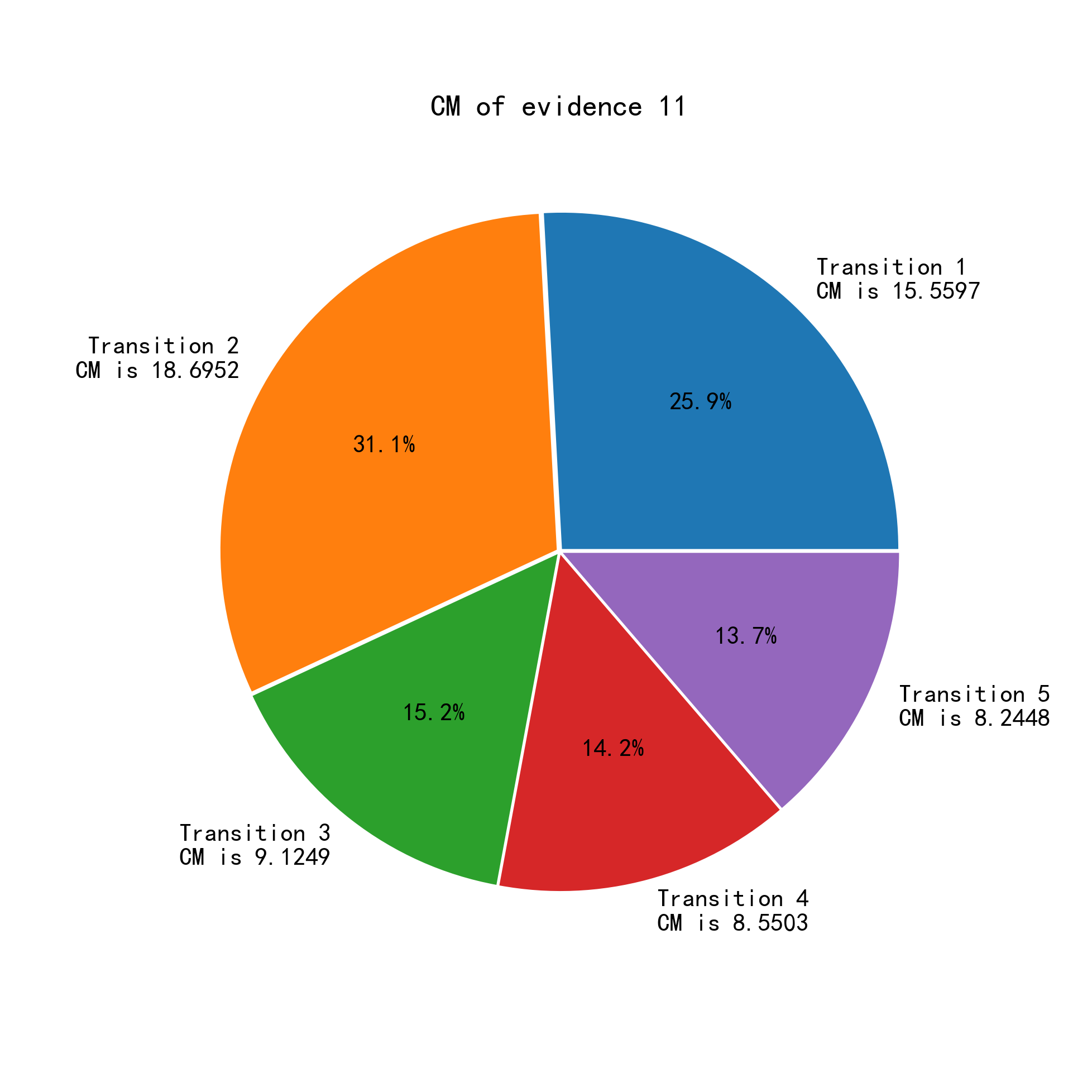}
	\caption{The proportion of different transitions in evidence 11}
	\label{6}
\end{figure}

\begin{figure}[h]
	\centering
	\includegraphics[scale=0.4]{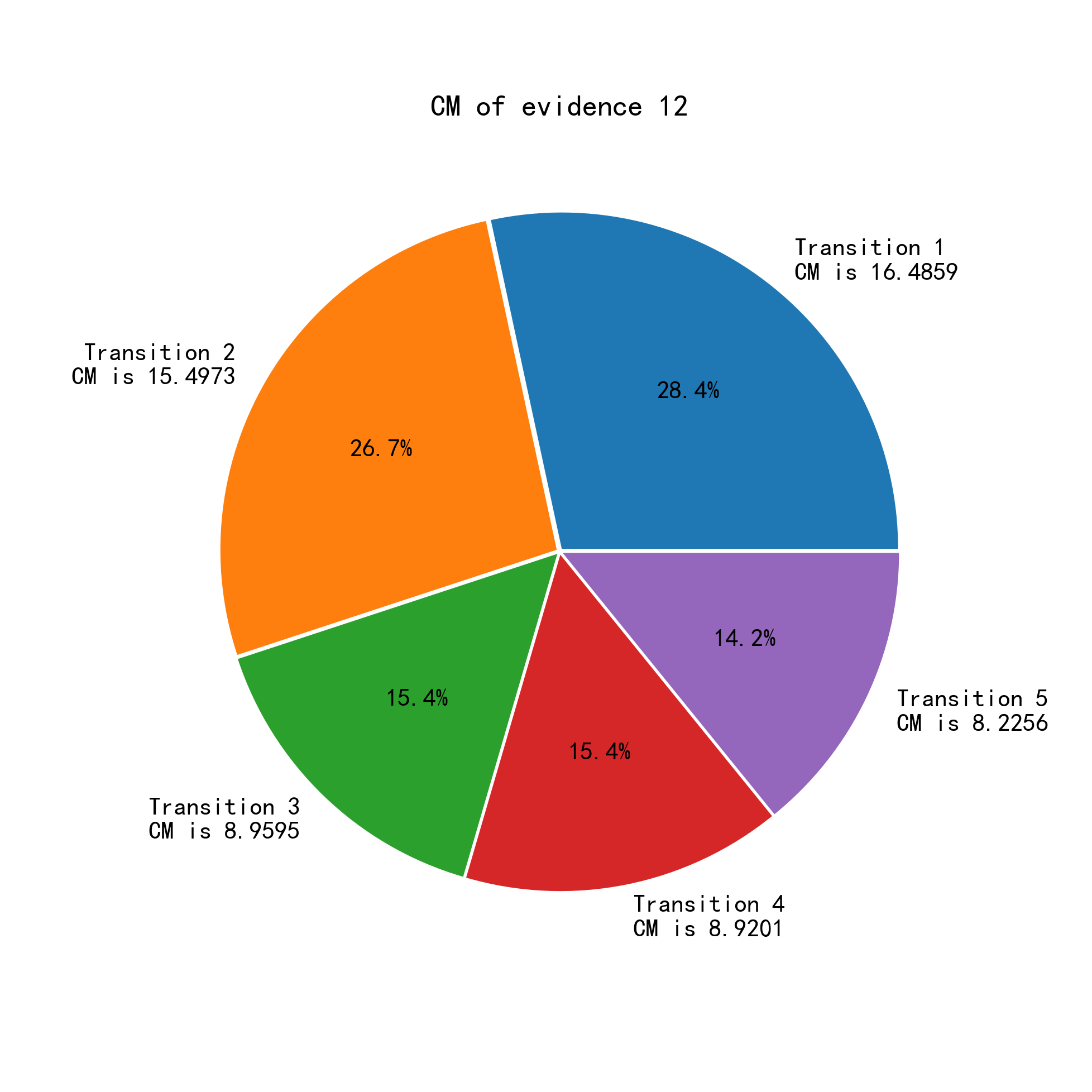}
	\caption{The proportion of different transitions in evidence 12}
	\label{7}
\end{figure}

\begin{figure}[h]
	\centering
	\includegraphics[scale=0.4]{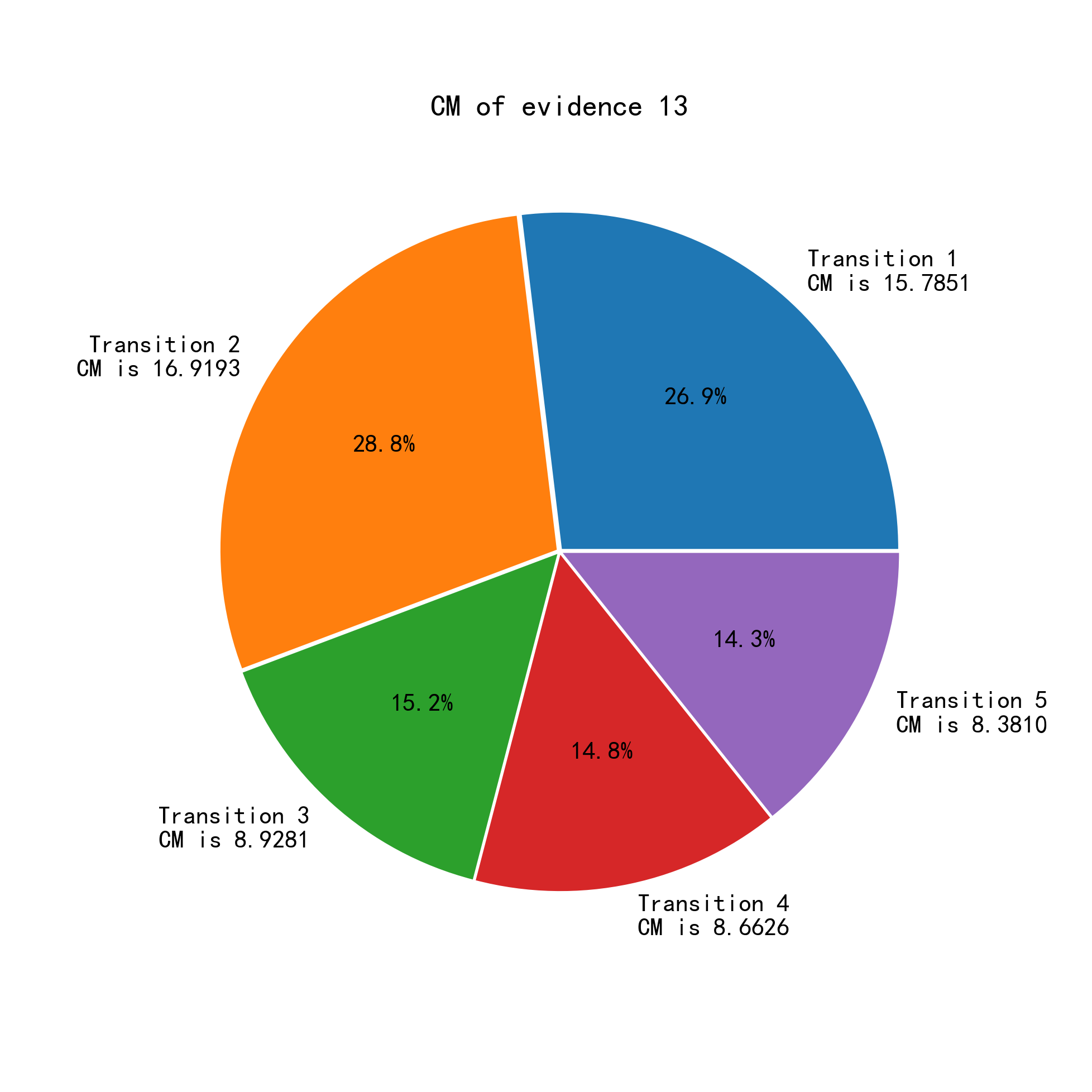}
	\caption{The proportion of different transitions in evidence 13}
	\label{8}
\end{figure}

\begin{figure}[h]
	\centering
	\includegraphics[scale=0.4]{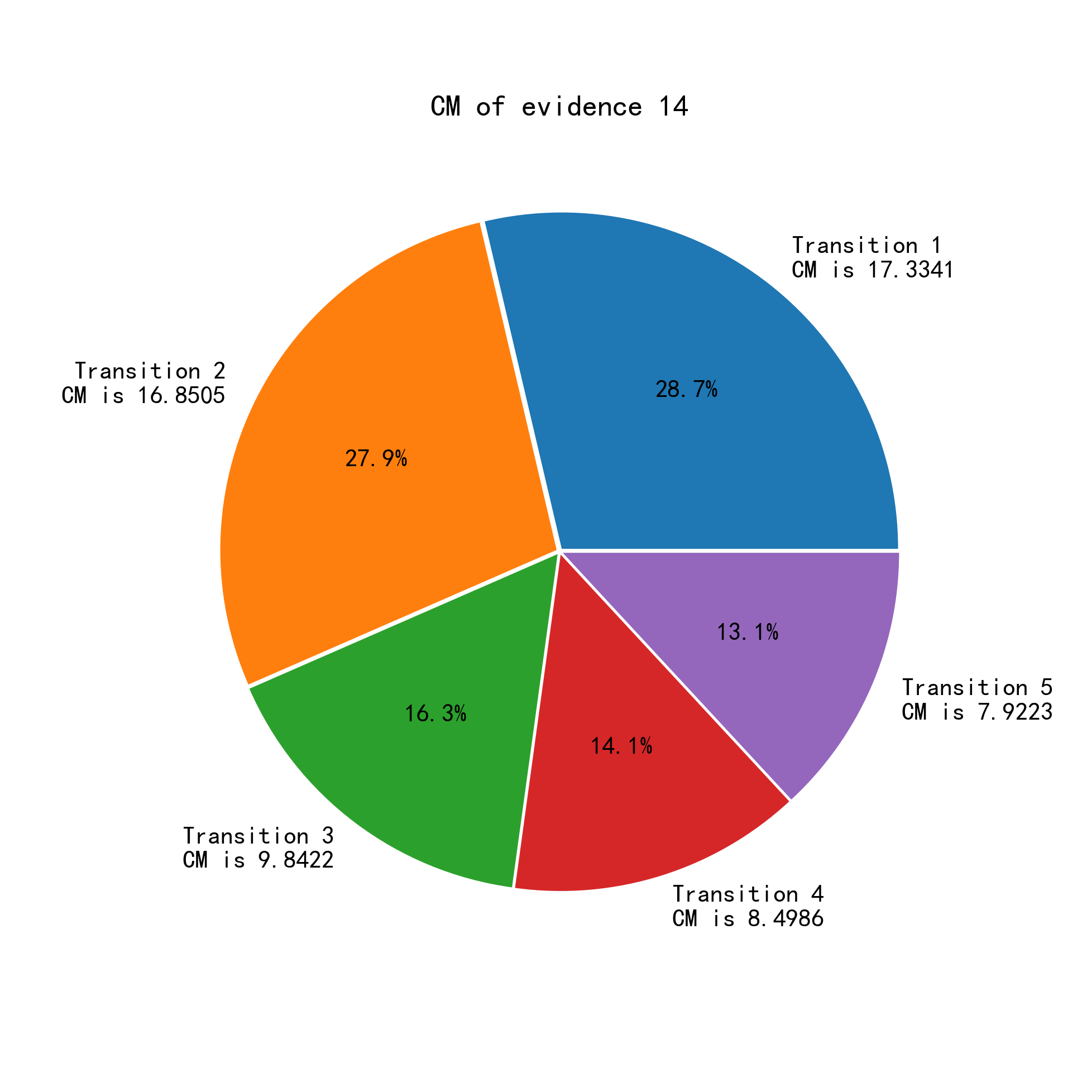}
	\caption{The proportion of different transitions in evidence 14}
	\label{sub5}
\end{figure}

Moreover, the CSs and CMs are obtained and presented in which CMs are given in Figure \ref{1}, \ref{2}, \ref{3}, \ref{4}, \ref{5}, \ref{6}, \ref{7}, \ref{8} and \ref{sub5} in the form of sector diagram. 

After obtaining the CMs for different CS, the modified evidences can be obtained through the procedure proposed before. And the results of modified evidences are given in Table \ref{ererer}. By analysing the figures produced based on the distribution of values of propositions, it can be summarised that the proportions of mass of different propositions are very similar which indicates that the effect of eliminating abnormal evidences of the step of normal distribution is sufficient.

And the mass of proposition $a$ and $b$ is the biggest in the propositions of evidences, which means the probability to observe this kind of transition is relatively small. On another side, the other propositions tend to transfer to state corresponding to the propositions which possess a higher mass. Therefore, if some propositions are the targets of other proposition to transfer to, then theses ones can be properly regarded as a much more credible objects to be put trust on. And that is what the CMs are exactly doing. The propositions with higher mass are allocated to CMs which owns a much higher weight in the whole body of evidences, which can further ensure a clear judgment about the true information contained in the evidences provided and is helpful to extract the most valuable information from uncertainty.

\begin{table*}[h]\footnotesize
	\centering
	\caption{The evidences after modification}
			\begin{tabular}{c c c c c c c c c c c c c c c c}\hline
				$Evidences$ & \multicolumn{5}{c}{$Values \ \ of \ \ propositions $}&$Evidences$ & \multicolumn{5}{c}{$Values \ \ of \ \ propositions $}\\\hline
				$Evidence_{1}$& $\{a\}$ & $\{b\}$ &$\{c\}$&$\{d\}$&$\{\emptyset\}$&$Evidence_{3}$& $\{a\}$ & $\{b\}$ &$\{c\}$&$\{d\}$&$\{\emptyset\}$ \\
				& $0.3287 $ & $0.4261 $ & $0.0570 $ & $0.0895 $ & $0.0986 $ && $0.2997 $ & $0.4717 $ & $0.0871 $ & $0.0628 $ & $0.0786 $\\
				$Evidence_{6}$& $\{a\}$ & $\{b\}$ &$\{c\}$&$\{d\}$&$\{\emptyset\}$&$Evidence_{9}$& $\{a\}$ & $\{b\}$ &$\{c\}$&$\{d\}$&$\{\emptyset\}$ \\
				& $0.3760 $ & $0.4277 $ & $0.0755 $ & $0.0532 $ & $0.0677 $ & & $0.3434 $ & $0.4306 $ & $0.0806 $ & $0.0891 $ & $0.0563 $\\
				$Evidence_{10}$& $\{a\}$ & $\{b\}$ &$\{c\}$&$\{d\}$&$\{\emptyset\}$&$Evidence_{11}$& $\{a\}$ & $\{b\}$ &$\{c\}$&$\{d\}$&$\{\emptyset\}$ \\
				& $0.3963 $ & $0.4254 $ & $0.0728 $ & $0.0660 $ & $0.0395 $ & & $0.3393 $ & $0.4743 $ & $0.0813 $ & $0.0586 $ & $0.0446 $\\
				$Evidence_{12}$& $\{a\}$ & $\{b\}$ &$\{c\}$&$\{d\}$&$\{\emptyset\}$&$Evidence_{13}$& $\{a\}$ & $\{b\}$ &$\{c\}$&$\{d\}$&$\{\emptyset\}$ \\
				& $0.4135 $ & $0.3681 $ & $0.0839 $ & $0.0823 $ & $0.0522 $ && $0.3716 $ & $0.4227 $ & $0.0799 $ & $0.0688 $ & $0.0569 $\\
				$Evidence_{14}$& $\{a\}$ & $\{b\}$ &$\{c\}$&$\{d\}$&$\{\emptyset\}$\\
				& $0.4126 $ & $0.3919 $ & $0.1080 $ & $0.0554 $ & $0.0321 $\\
				\hline
		\end{tabular}
	
	\label{ererer}
\end{table*}

\subsection{The combination of the MMGET}
After getting the modified evidences, the combination of the modified evidences can be given according to the definition of combination rule for MMGET. All of the results are listed in Table \ref{table8}.

\begin{table*}[h]\footnotesize
	\centering
	\caption{The comparison of results of combination using two methods}
	
			\begin{tabular}{c c c c c c}\hline
				$Proposed\ \ method$ & \multicolumn{5}{c}{$Values \ \ of \ \ propositions $}\\\hline
				$Propositions$& $\{a\}$ & $\{b\}$ &$\{c\}$&$\{d\}$&$\{\emptyset\}$\\\hline
				& $0.1923 $ & $0.8076 $ & $2.2972e-07 $ & $5.7334e-08 $ & $5.1628e-12 $\\\hline
				$Traditional\ \ method\ \ of\ \ combination$ & \multicolumn{5}{c}{$Values \ \ of \ \ propositions $}\\\hline
				$Propositions$& $\{a\}$ & $\{b\}$ &$\{c\}$&$\{d\}$&$\{\emptyset\}$ \\\hline
				 & $0.3380 $ & $0.6619 $ & $6.0952e-05 $ & $2.0272e-05 $ & $4.7355e-10 $\\
				\hline
		\end{tabular}
	\label{table8}
\end{table*}

It can be easily obtained that two kinds of method indicate the proposition $a$ and $b$ own the most two biggest probability to take place. However, for the detailed values of the corresponding propositions, there exists a distinct differences in the distribution of values of indicator in the probability of according propositions. When inspecting the situation of allocation of mass of the two propositions in evidences provided, it can be concluded that almost all of the values of proposition $b$ are relatively much bigger than those of proposition $a$, which indicates that the proposition or incident $b$ is expected to take place with a much bigger probability. Based on the phenomenon obtained from the conditions of evidences, it can be concluded that the results of combination are supposed to support the proposition $b$ by assigning a bigger mass to it. However, in the results obtained by traditional method, the distinction of the mass of proposition is not so obvious like the ones in the results of the proposed method, which means the proposed method has a better ability in indicating the actual situation the evidences describe and reduce uncertainty contained in the information from evidences. All in all, the proposed method performs much better in indicating the incident or proposition which is the most possible to happen.

\subsection{Certainty measure of FOD}
\begin{figure*}[h]
	\centering
	\includegraphics[scale=0.6]{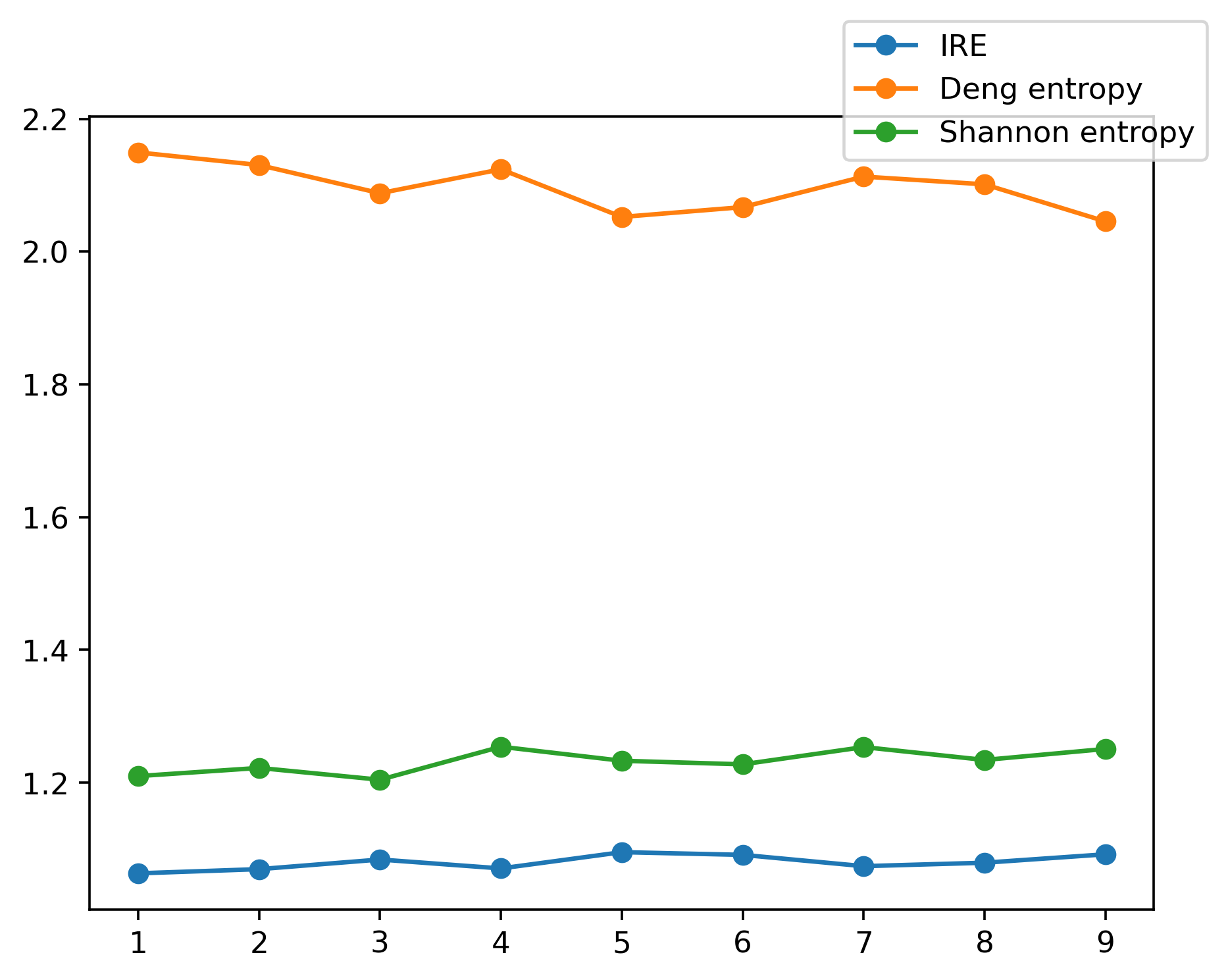}
	\caption{The fluctuation of values of 3 methods}
	\label{66}
\end{figure*}

From the definition of CRE and utilizing the data of TP, the specific values of CRE are obtained. Besides, the results of CRE are compared with the corresponding values of two kinds of entropies, namely Deng entropy and Shannon entropy which are listed in Table \ref{table12}. And the fluctuation of the values are also given in the Figure \ref{66} to help distinguish the difference of trend of corresponding entropies.

\begin{table*}[h]\footnotesize
	\centering
	\caption{The comparison of results of uncertainty measure using three methods}
		\begin{tabular}{c c c c c c c c c c c c c c c c}\hline
			$Evidences$ & $Values \ \ of \ \ IRE$&$Deng\ \ entropy$ & $Shannon\ \ entropy$\\\hline
			$Evidence_{1}$&$1.0634$ &$2.1494$ &$1.2099$ \\
			$Evidence_{3}$&$1.0697$ &$2.1304$ &$1.2222$ \\
			$Evidence_{6}$&$1.0842$ &$2.0879$ &$1.2044$ \\
			$Evidence_{9}$&$1.0708$ &$2.1240$ &$1.2541$ \\
			$Evidence_{10}$&$1.0953$ &$2.0522$ &$1.2330$ \\
			$Evidence_{11}$&$1.0912$ &$2.0670$ &$1.2276$ \\
			$Evidence_{12}$&$1.0742$ &$2.1130$ &$1.2536$ \\
			$Evidence_{13}$&$1.0794$ &$2.1013$ &$1.2342$ \\
			$Evidence_{14}$&$1.0921$ &$2.0454$ &$1.2508$ \\
			\hline
		\end{tabular}
	\label{table12}
\end{table*}

In fact, the IRE take the TPs and the cardinality of the propositions in the process of transition into consideration which can be regarded as crucial indicators of the uncertainty of measure. One of the most important reasons to considers the TP is that if the condition of one piece of evidence can be regarded as very certain, then the effect of indicating one or two incident to take place is supposed to be very obvious, which also means the propositions are expected to have a much bigger mass than the rest of propositions. Then, the TPs of the rest of propositions are set at a very high level, which leads a higher value of IRE at the same time. On the contrary, if there exists no proposition whose value is prominently higher than others, then the TPs will retain at a relatively low standard. Besides, another reason to take the subtraction of the cardinality of propositions is that when a multiple proposition transits to a state whose corresponding propositions' cardinality is smaller than the original one, then the increment of degree of certainty could be exaggerated appropriately because this kind of transition cause the degree of uncertainty to be reduced. All the features contribute that when the value of IRE rises, the conditions described by the evidences are more certain and when the value of IRE decreases, the whole FOD tends to become more and more uncertain.

By analysing the trend of values of the methods, some interesting points can be harvested. First, the trends of IRE and Deng entropy are completely opposite. For Deng entropy, the rise of values of entropy means that the whole FOD tends to be more uncertain and when it decreases, the FOD becomes more certain. And according to the definition of the IRE, a more certain condition is indicated when the value rises. On the opposite, if the value of IRE is lowered, the whole FOD tends become chaotic and uncertain. Due to the reverse trend of the two methods, it can be reasonably concluded that the effectiveness of the two method is almost the same. A tiny difference is that gap between different evidences. The ones presented by IRE are less obvious than Deng entropy, considering the small differences of evidences, it is reasonable to allocate similar values of results instead of giving much more divergent values. Second, compared with Shannon entropy, it is intuitive to think that the trend of the two methods are opposite, most of the part of the broken line accords with the conclusion. However, the part of $1\ \ to\ \ 2$, $5\ \ to\ \ 6$ and $8\ \ to\ \ 9$ do not conform the conclusion made. For the part of $1\ \ to\ \ 2$, the mass of proposition $b$ increases and the increment is more than the decrement of proposition $a$, which indicates that the effect of indicator of probability of specific propositions is strengthened. So the whole frame is supposed to become more certain instead of having a higher degree of uncertainty. And with respect to the part of $5\ \ to\ \ 6$, the increment of the proposition $b$ is much less than the decrement of proposition $a$ so that the effectiveness of indicator of the real situation is alleviated to some extent. Therefore, for the Shannon entropy, the value is supposed to be raised. In the last, for the part of $8\ \ to\ \ 9$, the sum of proposition $a$ and $b$ and the value of $a$ increases, there is no reason to reckon the FOD become uncertain. All in all, the CRE owns all the advantages of Deng entropy and could present the credential degree of evidences directly, which simplify the process to allocate a weight to corresponding evidences in disposing information without extra operations.

\section{Conclusion}

In this paper, a Markov model is introduced into the GET, which enables the improved version of GET to better adapt to actual situations and an open world. The proposed model provides a completely new vision on the solution of similarity measure, certainty measure, combination of evidences and distance measure, which can offers a much better accuracy and intuitive results than other methods bu utilizing every details of evidences. The Markov GET can be regarded as an optimal choice in handling information given in the form of evidence and a complete solution to obtain kinds of indexes of evidences in a reasonable manner.

\section*{Acknowledgment}
The authors greatly appreciate the reviewers’ suggestions and the editor’s
encouragement. This research is supported by the National Natural Science
Foundation of China (No. 62003280).

%
\IEEEpeerreviewmaketitle



%

\bibliographystyle{IEEEtran}
\bibliography{cite}


%

%
%
%




\end{document}